\documentclass[letterpaper]{article} 
\usepackage{aaai23}  
\usepackage{times}  
\usepackage{helvet}  
\usepackage{courier}  
\usepackage[hyphens]{url}  
\usepackage{graphicx} 
\usepackage{natbib}  
\usepackage{caption} 
\usepackage{xcolor}

\usepackage{wrapfig}
\usepackage{algorithm}
\usepackage{algpseudocode}
\usepackage{amssymb} 
\usepackage{mathtools}
\usepackage{subcaption}

\usepackage{newfloat}
\usepackage{listings}
\DeclareCaptionStyle{ruled}{labelfont=normalfont,labelsep=colon,strut=off} 
\floatstyle{ruled}
\newfloat{listing}{tb}{lst}{}
\floatname{listing}{Listing}
\title{Toward Robust Uncertainty Estimation with Random Activation Functions}
\author {
    Yana Stoyanova, 
    Soroush Ghandi, 
    Maryam Tavakol 
}
\affiliations {
    Eindhoven University of Technology, Eindhoven, The Netherlands\\
    y.stoyanova@student.tue.nl,
    s.ghandi@tue.nl,
    m.tavakol@tue.nl
}

\usepackage{bibentry}

\begin{document}

\maketitle

\begin{abstract}
Deep neural networks are in the limelight of machine learning with their excellent performance in many data-driven applications. However, they can lead to inaccurate predictions when queried in out-of-distribution data points, which can have detrimental effects especially in sensitive domains, such as healthcare and transportation, where erroneous predictions can be very costly and/or dangerous. Subsequently, quantifying the uncertainty of the output of a neural network is often leveraged to evaluate the confidence of its predictions, and ensemble models have proved to be effective in measuring the uncertainty by utilizing the variance of predictions over a pool of models. In this paper, we propose a novel approach for uncertainty quantification via ensembles, called \textit{Random Activation Functions (RAFs) Ensemble}, that aims at improving the ensemble diversity toward a more robust estimation, by accommodating each neural network with a different (random) activation function. Extensive empirical study demonstrates that RAFs Ensemble outperforms state-of-the-art ensemble uncertainty quantification methods on both synthetic and real-world datasets in a series of regression tasks.
\end{abstract}

\section{Introduction}
Recent advances in deep neural networks have demonstrated remarkable performance in a wide variety of applications, ranging from recommendation systems and improving user experience to natural language processing and speech recognition
\cite{Abiodun2018StateoftheartIA}. Nevertheless, blindly relying on the outcome of these models  can have harmful effects, especially in high-stake domains such as healthcare and autonomous driving, as models can provide inaccurate predictions when queried in out-of-distribution data points 
\cite{Amodei2016ConcretePI}. 
Consequently, correctly quantifying the uncertainty of models' predictions is an admissible mechanism to 
distinguish where a model can or cannot be trusted, and thus, increases the transparency of models about their capabilities and limitations \cite{Abdar2021ARO}. Uncertainty Quantification (UQ) is important for a variety of reasons. For instance, in order to preserve the model’s credibility, it is essential to report and communicate the encountered uncertainties regularly \cite{Volodina2021TheReproducibility}. Additionally, models' predictions are inevitably uncertain in most cases, which has to be addressed to increase their transparency, trustworthiness, and reliability.

In the machine learning literature, uncertainty is usually decomposed into two different types, namely aleatoric uncertainty and epistemic uncertainty \cite{Kiureghian2009AleatoryOE}. \textit{Aleatoric} uncertainty, aka data uncertainty, refers to the inherent uncertainty that stems from the data itself, e.g., noise. On the other hand, \textit{epistemic} uncertainty, also called model uncertainty, is the type of uncertainty that occurs due to the lack of sufficient data. While data uncertainty \textit{cannot} be alleviated, model uncertainty can be addressed by e.g., acquiring more data. Let $\boldsymbol{\sigma}_a^2$ and $\boldsymbol{\sigma}_e^2$ denote the aleatoric and epistemic uncertainties, respectively. Since the distinction between the two is imprecise to some degree \cite{Sullivan2015IntroductionTU}, we focus on the predictive (total) uncertainty, which is defined as the sum of the two
%
\begin{equation}
\label{eq:unc}
    \boldsymbol{\sigma}_p^2 = \boldsymbol{\sigma}_a^2 + \boldsymbol{\sigma}_e^2.
\end{equation}
%
%
Accordingly, the approaches developed for uncertainty estimation can be categorized into three groups: Bayesian UQ methods, ensemble UQ methods, and a combination of both, i.e., Bayesian ensemble UQ  \cite{Abdar2021ARO}. In this paper, we focus on ensemble UQ techniques, either Bayesian or non-Bayesian, as this  group is less explored compared to the solely Bayesian techniques. An ensemble model aggregates the predictions of multiple individual base-learners (or ensemble members), which in our case  are neural networks (NNs), and the empirical variance of their predictions gives an approximate measure of uncertainty. The idea behind this heuristic is highly intuitive: the more the base-learners disagree on the outcome, the more uncertain they are. Therefore, the goal of ensemble members is to have a great level of disagreement (variability) in the areas where little or no data is available, and to have a high level of agreement in regions with abundance of data \cite{Pearce2018UncertaintyIN}. 

In this paper, we propose a novel method, called \textit{Random Activation Functions Ensemble (RAFs Ensemble)}, for a more robust uncertainty estimation in (deep) neural networks. RAFs Ensemble is developed on top of Anchored Ensemble technique, proposed by \cite{Pearce2018UncertaintyIN}, however, instead of initializing each NN member in the ensemble with the same activation function, the NNs in RAFs Ensemble are accommodated with different (random) activation functions in the hidden layers. This simple, yet crucial, modification greatly improves the overall diversity of the ensemble, which is one of the most important components in forming a successful ensemble. We empirically show that RAFs Ensemble provides high quality uncertainty estimates compared to five state-of-the-art ensemble methods, that is Deep Ensemble \cite{Lakshminarayanan2017SimpleAS}, Neural Tangent Kernel Gaussian Process Parameter Ensemble \cite{He2020BayesianDE}, Anchored Ensemble \cite{Pearce2018UncertaintyIN}, Bootstrapped Ensemble of NNs Coupled with Random Priors  \cite{Osband2018RandomizedPF}, and Hyperdeep Ensemble \cite{Wenzel2020HyperparameterEF}. The comparisons are performed in a wide range of regression tasks on both synthetic and real-world datasets in terms of negative log-likelihood and root mean squared error. 

\section{Related Work}

Uncertainty Quantification (UQ) is an active field of research and various methods have been proposed to efficiently estimate the uncertainty of machine learning models (see \citeauthor{Abdar2021ARO} \citeyear{Abdar2021ARO} for an extensive overview).
While most research focuses on Bayesian deep learning \cite{Srivastava2014DropoutAS, Blundell2015WeightUI, Sensoy2018EvidentialDL, Fan2020BayesianAM, Jrvenp2020BatchSA, Charpentier2020PosteriorNU}, deep ensemble methods, which benefit from the advantages of both deep learning and ensemble learning, have been recently leveraged for empirical uncertainty quantification \cite{Egele2021AutoDEUQAD, Hoffmann2021UncertaintyQB, Brown2020UncertaintyQI, Althoff2021UncertaintyQF}. Although Bayesian UQ methods have solid theoretical foundation, they often require significant changes to the training procedure and are computationally expensive compared to non-Bayesian techniques such as ensembles
\cite{Egele2021AutoDEUQAD, Rahaman2021UncertaintyQA, Lakshminarayanan2017SimpleAS}. 
 
\citeauthor{Lakshminarayanan2017SimpleAS}~\shortcite{Lakshminarayanan2017SimpleAS} are among the first to challenge  Bayesian UQ methods by proposing Deep Ensemble, a simple and scalable technique, 
that demonstrates superb empirical performance on a variety of datasets. However, one of the challenges of ensemble techniques when quantifying uncertainty is that they tend to give overconfident predictions \cite{Amodei2016ConcretePI}. 
To address this, \citeauthor{Pearce2018UncertaintyIN}~\shortcite{Pearce2018UncertaintyIN} propose to also regularize the model's parameters w.r.t. the initialization values, instead of zero, leading to Anchored Ensembles, which additionally allows for performing Bayesian inference in NNs. 
\citeauthor{He2020BayesianDE}~\shortcite{He2020BayesianDE} relate Deep Ensembles to Bayesian inference using neural tangent kernels. Their method, i.e., Neural Tangent Kernel Gaussian Process Parameter Ensemble (NTKGP-param), trains all layers of a finite width NN, obtaining an exact posterior interpretation in the infinite width limit with neural tangent kernel parameterization and squared error loss. They prove that NTKGP-param is always more conservative than Deep Ensemble, yet, its advantages are generally not clear in practice.


A prominent advance to the Bayesian ensemble UQ methods is the bootstrapped ensemble of NNs coupled with random priors, proposed by \cite{Osband2018RandomizedPF}, in which, 
the random prior function and neural models share an input and a summed output, but the networks are the only trainable parts, while the random prior remains untrained throughout the whole process. 
Furthermore, \citeauthor{Wenzel2020HyperparameterEF}~\shortcite{Wenzel2020HyperparameterEF}
exploit an additional source of randomness in ensembles by designing ensembles not only over weights, but also over hyperparameters. Their method, called Hyperdeep Ensemble, demonstrates high accuracy for a number of different classification tasks. Nevertheless, despite the recent contributions in ensemble UQ methods, the research in this direction still needs further advancement.

\section{Toward Robust Uncertainty Estimation}
\subsection{Preliminaries}
Following the notations of \cite{Lakshminarayanan2017SimpleAS},
let $S_{train}$ be a training dataset  consisting of $n$ independently and identically drawn (i.i.d.) data points, $S_{train} = \{\boldsymbol{x}_i,y_i\}_{i=1}^n$, where $\boldsymbol{x}_i \in \mathbb{R}^d$ denotes  a $d$-dimensional feature vector and $y_i \in \mathbb{R}$ is a scalar output. Similarly, $S_{test}$
indicates the test set.
Subsequently, $X$ represents the design matrix and $\boldsymbol{y}$ indicates the output vector, where $(S_{train}.X,  S_{train}.\boldsymbol{y})$ and $(S_{test}.X,  S_{test}.\boldsymbol{y})$ represent the train and test sets, respectively.
Without the loss of generality, we consider the regression tasks of the form
\begin{equation*}
    \boldsymbol{y} = f(X) + \epsilon, 
\end{equation*}
where $\epsilon$ is a normally distributed constant noise, i.e., $\epsilon \sim \mathcal{N}(0, \boldsymbol{\sigma}_a^2)$, and is assumed to be known. 
The goal is hence to quantify the predictive uncertainty $\boldsymbol{\sigma}_p^2$ associated with $S_{test}.\boldsymbol{y}$, while optimizing $f$ on the training data. 

We adapt the regularized loss function from the Anchored Ensemble technique~\cite{Pearce2018UncertaintyIN}, in which, the regularization of the models' parameters are carried out w.r.t. their initialization values instead of zero. Consequentially, given $\boldsymbol{\theta}_j$ as the parameters of the $j$\textsubscript{th} base-learner, the objective function is as follows
\begin{equation}
\label{anchor}
    \mathcal{L}(\boldsymbol{\theta}_j) = \frac{1}{n}||\boldsymbol{y}-\boldsymbol{\hat{y}}_j||_2^2+\frac{1}{n}||\Gamma^{1/2}(\boldsymbol{\theta}_j - \boldsymbol{\theta}_{0,j})||_2^2,
\end{equation}
where $\boldsymbol{\theta}_{0,j}$ is derived from the prior distribution, $\boldsymbol{\theta}_{0,j} \sim \mathcal{N}(\boldsymbol{\mu}_{0}, \Sigma_{0})$, and $\Gamma$ is the regularization matrix. Furthermore, minimizing this objective allows for performing Bayesian inference in NNs. However, this technique only models the epistemic uncertainty, while aleatoric uncertainty is assumed to be constant \cite{Pearce2018UncertaintyIN}, which is a limitation, as it is not always possible to distinguish the different origins or types of uncertainty in practice (see Equation~\ref{eq:unc}). 

Therefore, in this paper, we aim at enhancing the performance of the ensemble toward a more robust uncertainty estimation. The literature suggests that diversifying the ensembles is effective in improving their predictive performance both theoretically and empirically \cite{Zhou2012EnsembleMF, Zhang2012EnsembleML,Hansen1990NeuralNE, Krogh1994NeuralNE}.
Ideally, diversity is achieved when the predictions made by each model in the ensemble are independent and uncorrelated. However, generating diverse ensemble members is not a straightforward task. The main impediment is the fact that each neural network is trained on the same training data to solve the same problem, which usually results in a high correlation among the individual base-learners \cite{Zhou2012EnsembleMF}. In the subsequent section, we introduce a simple technique to efficiently improve the  overall diversity of the ensemble for a more reliable uncertainty quantification.


\subsection{RAFs Ensemble}
In this section, we present Random Activation Functions (RAFs) Ensemble for uncertainty estimation, which can be extended to all ensemble methods in terms of methodological modification.
When a (Bayesian) ensemble is leveraged to estimate the uncertainty of a deep neural network model, we propose to increase the diversity of predictions among the ensemble members using varied activation functions (AFs), in addition to the random initialization of the parameters. To do so, instead of initializing the neural networks with the same activation function, each NN is accommodated with a different (random) activation function. Subsequently, distinct activation functions account for different non-linear properties introduced to each ensemble member, therewith improving the overall diversity of the ensemble. 

As mentioned previously, the ensemble diversity is one of the most important building blocks when it comes to creating a successful ensemble \cite{Hansen1990NeuralNE}. Hence, it might be preferable to combine the predictions of top-performing base-learners with the predictions of weaker ones \cite{Zhou2012EnsembleMF}. Otherwise, stacking only strong models will likely result in a poor ensemble as the predictions made by the models will be highly correlated, and thus, the ensemble diversity will be greatly limited. Therefore, the choice of activation functions should be motivated purely by their variability and not their appropriateness for the task at-hand.

Let $\boldsymbol{\mu}_0$ be the prior means, $\Sigma_0$ be the prior covariance, $\boldsymbol{\hat{\sigma}}_a^2$ 
be an estimate of data noise, $m$ denote the number of base-learners, and $NN_j$ indicate the $j_\text{th}$ member, the entire procedure for both training and prediction is summarized in Algorithm \ref{alg:rafs}. In this algorithm, a regularization matrix is first created and a set of activation functions is defined  (line 1-2). Then, the NNs in the ensemble are trained to minimize the loss function defined in Equation \ref{anchor} with stochastic gradient descent, using arbitrary optimizer and no early stopping (line 3-13). Note that if the size of the ensemble $m$ is smaller or equal to the cardinality of the AFs set $k$, then each NN is trained with a different activation function, and with random functions from the set, otherwise (line 7-11). Consequently, predictions are made with each ensemble member (line 14-16), which are then averaged and an estimate of the predictive uncertainty is computed (line 17-19). 



\begin{algorithm}[t]
\caption{RAFs Ensemble}\label{alg:rafs}
\textbf{Input:} $S_{train}, S_{test},$ priors $\boldsymbol{\mu}_0$ and $\Sigma_0$, $m$, $\boldsymbol{\hat{\sigma}}_a^2$\\
\textbf{Output:} Estimate of predictive mean  $\boldsymbol{\hat{y}}$ and variance  $\boldsymbol{\hat{\sigma}}_p^2$
\begin{algorithmic}[1]
\State $\Gamma \leftarrow \boldsymbol{\hat{\sigma}}_a^2 \Sigma_0^{-1}$   \Comment{Regularization matrix}
\State $\mathbb{A}\leftarrow \{a_1,\ldots,a_k\}$ \Comment{Set of $k$ AFs}
\For{$\hspace{1mm} j$ in $1:m$}    \Comment{Train the ensemble}
\State Create $NN_j$ with $\boldsymbol{\theta}_{j,0}\leftarrow \mathcal{N}(\boldsymbol{\mu}_0, \Sigma_0)$
\If{$ j \leq k$}
    \State $\alpha_j=a_j$
\Else
    \State $\alpha_j\leftarrow$ Randomly selected from $\mathbb{A}$
\EndIf
\State $NN_j.train(S_{train},\Gamma, \boldsymbol{\theta}_{j,0}, \alpha_j)$ using loss in Eq. \ref{anchor}
\EndFor
\For{$\hspace{1mm} j$ in $1:m$}   \Comment{Predict with the ensemble}
\State $\boldsymbol{\hat{y}}_j = NN_j.predict(S_{test}.X)$
\EndFor
\State $\boldsymbol{\hat{y}} = \frac{1}{m}\displaystyle\sum_{j=1}^{\mathclap{m}} \boldsymbol{\hat{y}}_j $ \Comment{Mean predictions}
\State $\boldsymbol{\hat{\sigma}}_e^2 = \frac{1}{m-1} \displaystyle\sum_{j=1}^{\mathclap{m}} (\boldsymbol{\hat{y}}_j - \boldsymbol{\hat{y}})^2$   \Comment{Epistemic variance}
\State $\boldsymbol{\hat{\sigma}}_p^2 = \boldsymbol{\hat{\sigma}}_e^2 + \boldsymbol{\hat{\sigma}}_a^2$ \Comment{Total variance Eq. \ref{eq:unc}}
\State \Return $\boldsymbol{\hat{y}}, \boldsymbol{\hat{\sigma}}_p^2$ 
\end{algorithmic}
\end{algorithm}

\section{Empirical Study}

\subsection{Experimental Setups}
In the experiments, the base-learners of RAFs Ensemble are multilayer perceptrons that consist of one hidden layer of 100 neurons. The ensemble size $m$ is set to five. This is standard for the implementations of all methods in this paper, as $m=5$ proved to be empirically sufficient for obtaining predictive uncertainty estimates in the experiments. 
In addition, we choose a set of seven activation functions which is comprised of (i) Gaussian Error Linear Unit (GELU) \cite{Hendrycks2016GaussianEL}, (ii) Softsign \cite{Turian2009QuadraticFA}, (iii) Swish \cite{Ramachandran2018SearchingFA}, (iv) Scaled Exponential Linear Unit (SELU) \cite{Klambauer2017SelfNormalizingNN}, (v) hyperbolic tangent (tanh), (vi) error activation function, and (vii) linear (identity) activation function.
%
%
Furthermore, the number of testing samples is set to be always larger than the number of training points $n$ to detail the uncertainty. Moreover, to account for epistemic uncertainty, the synthetic testing feature vectors $\boldsymbol{x}\in S_{test}$ range over wider intervals compared to $\boldsymbol{x}\in S_{train}$ and both are sampled uniformly at random. 
The code is available at \url{https://github.com/YanasGH/RAFs_code} for reproducibility.

\subsubsection{Baselines.}We include five state-of-the-art methods as baselines for empirical comparison with RAFs Ensemble  as follows. (i) DE \cite{Lakshminarayanan2017SimpleAS}, (ii) AE \cite{Pearce2018UncertaintyIN}, (iii) HDE \cite{Wenzel2020HyperparameterEF}, (iv) RP-param \cite{Osband2018RandomizedPF}, and (v) NTKGP-param \cite{He2020BayesianDE}, on both synthetic and real-world datasets with different dimensionalities (see the Technical appendix for a detailed overview). To ensure fair comparison between the UQ techniques, roughly the same amount of time has been put into hyperparameter tuning for each method. 

\subsubsection{Synthetic Data.} 
We generate multiple synthetic datasets that fall into four  categories: physical models (PM), many local minima (MLM), trigonometric (T), and others (O). Each set in the PM category is generated from a physical mathematical model, such that all values in $S_{train}$ and $S_{test}$ are achievable in the real world. Generally, the PM datasets have complex modeling dynamics and can be characterized as having predominant epistemic uncertainty due to the considerably wider testing sampling regions by design. Similarly, the MLM data, generated from functions with many local minima, are also designed so that the model uncertainty is higher than the aleatoric one. These datasets are usually hard to approximate due to their inherent high-nonlinearity and multimodality. Another category with higher epistemic uncertainty is trigonometric, such as
data generated by \cite{He2020BayesianDE} and \cite{Forrester2008EngineeringDV}, where the training data is partitioned into two equal-sized clusters in order to detail uncertainty on out-of-distribution data (see Figure \ref{fig:xsinx}). 
In contrast, the predominant type of uncertainty in the O category is aleatoric. This category includes datasets generated from various functions such as rational and product integrand functions. It is distinguished from the rest of the categories by its high interaction effects. The dimensionality of all datasets can range from one to ten and we consider two datasets per dimension, thus, the total number of synthetic data is 20. More detail on how the data is created can be found in the Technical appendix.

\subsubsection{Real-world Data.}
Additionally, we use five real-world datasets for evaluation: Boston housing, Abalone shells \cite{Abalone1994}, Naval propulsion plant \cite{Coraddu2013Machine}, Forest fire \cite{Cortez2007ADM}, and Parkinson's disease dataset \cite{Little2007ExploitingNR}. To account for aleatoric uncertainty (some) context factors are disregarded, such that this type of uncertainty is characteristically high (see Technical appendix for more details).

\subsubsection{Evaluation Criteria.}
\label{eval_crit}
We employ two evaluation criteria  to gauge the overall performance of the trained models, namely calibration and robustness to the distribution shift. Both measures are inspired by the practical applications of NNs, as generally there is no theoretical evidence for evaluating uncertainty estimates \cite{Abdar2021ARO}. Calibration is defined as the analytical process of adjusting the inputs with the purpose of making the model to predict the actual observations as precisely as possible \cite{Bijak2021UncertaintyQM}. The quality of calibration can be measured by proper scoring rules such as negative log-likelihood (NLL). NLL is a common choice when it comes to evaluating UQ estimates, as it depends on predictive uncertainty \cite{Lakshminarayanan2017SimpleAS}. Additionally, due to its practical applicability in a wide spectrum of regression tasks, root mean squared error (RMSE) is measured, although it does not depend on the estimated uncertainty \cite{Lakshminarayanan2017SimpleAS}, but serves as a proxy and a secondary assessor of the performance. Moreover, to measure the robustness/generalization of methods to distributional shift, we test the models in out-of-distribution settings, such as the synthetic datasets by \cite{Forrester2008EngineeringDV,He2020BayesianDE}. 



\begin{figure}[t]
\begin{subfigure}{.15\textwidth}
  \centering
  \includegraphics[width=\linewidth]{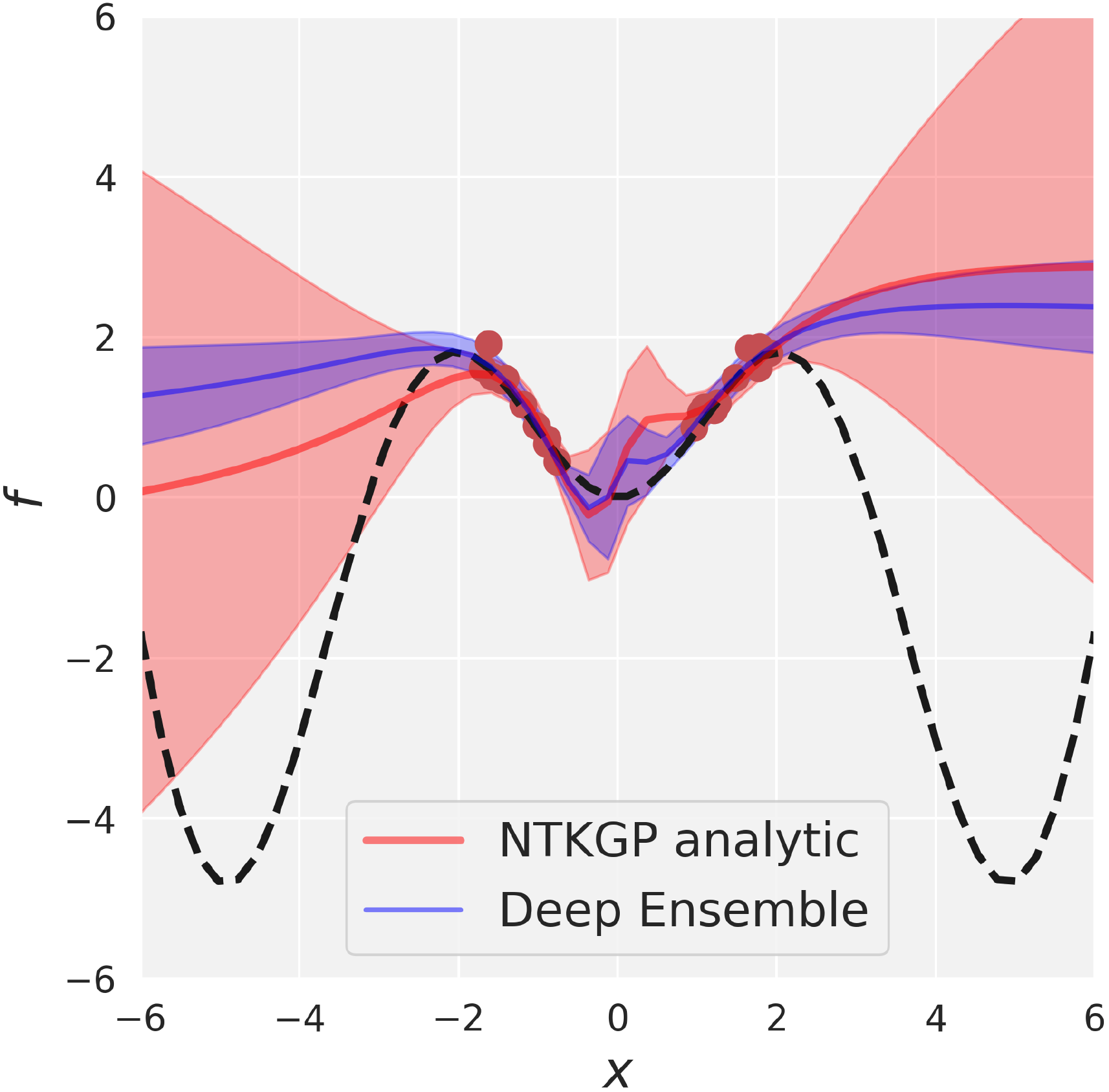}
  \caption{DE}
  \label{fig:xsinxDE}
\end{subfigure}%
\begin{subfigure}{.15\textwidth}
  \centering
  \includegraphics[width=\linewidth]{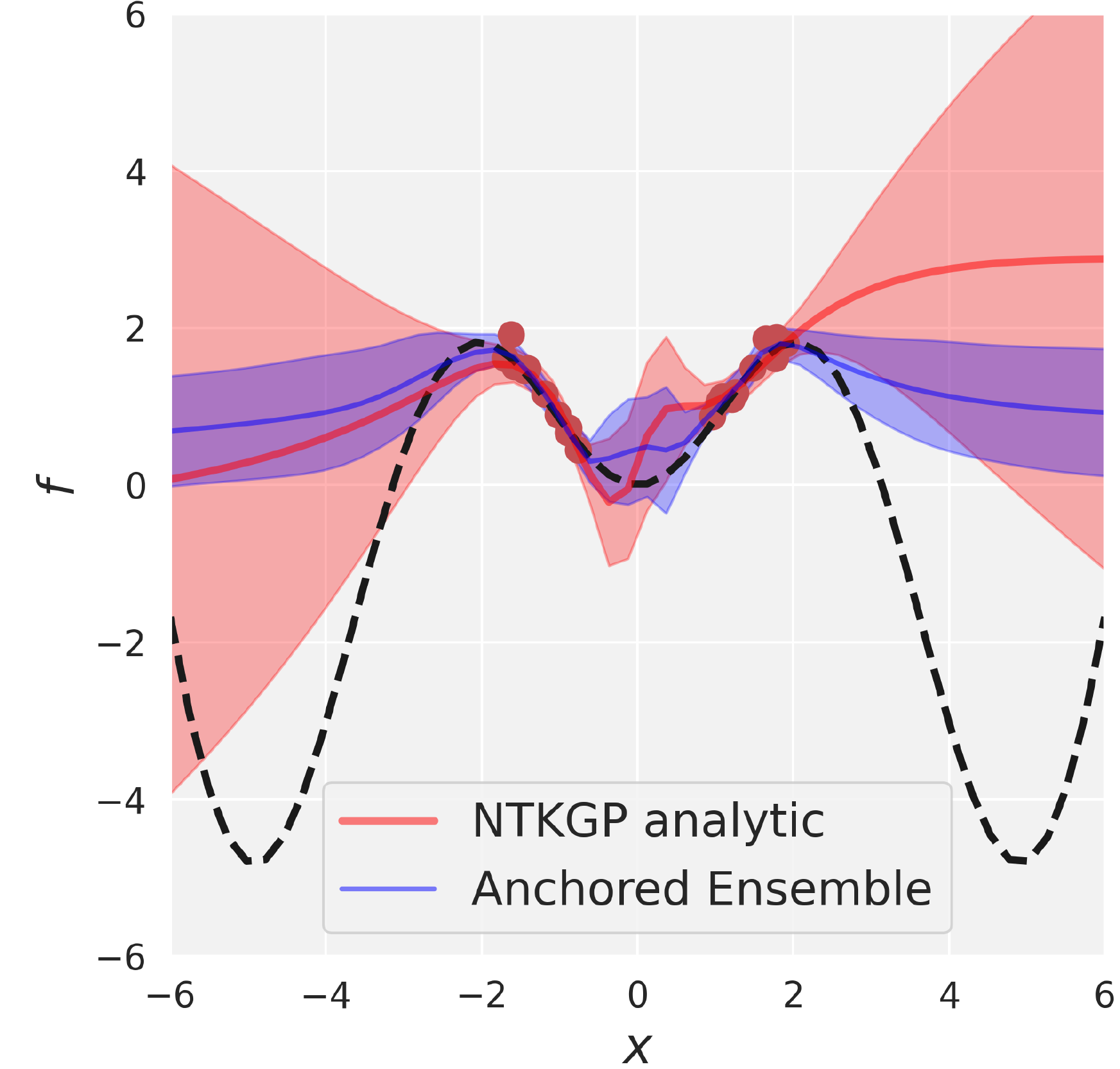}
  \caption{AE}
  \label{fig:xsinxAE}
\end{subfigure}
\begin{subfigure}{.15\textwidth}
  \centering
  \includegraphics[width=\linewidth]{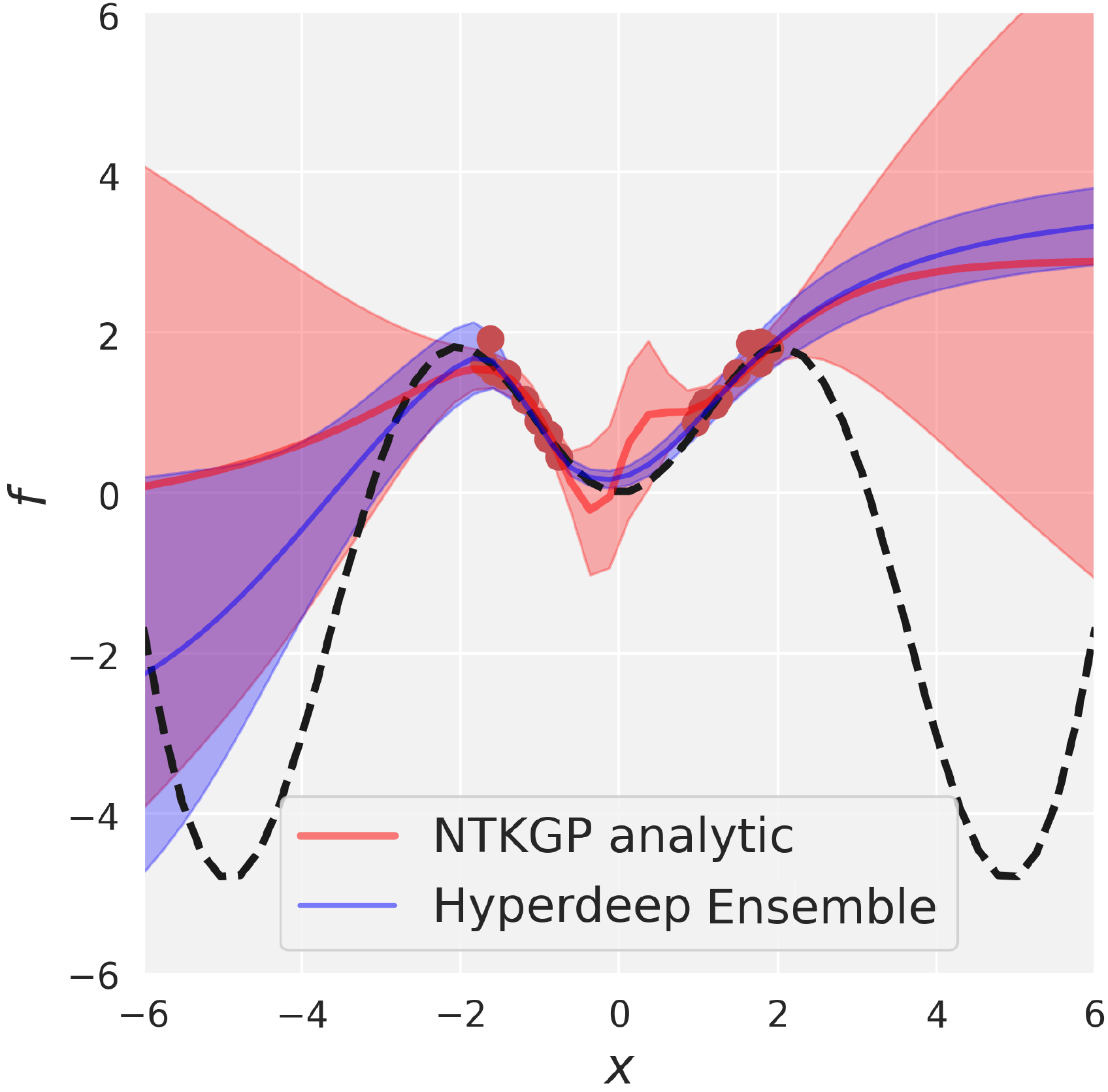}
  \caption{HDE}
  \label{fig:xsinxHDE}
\end{subfigure}

\begin{subfigure}{.15\textwidth}
  \centering
  \includegraphics[width=\linewidth]{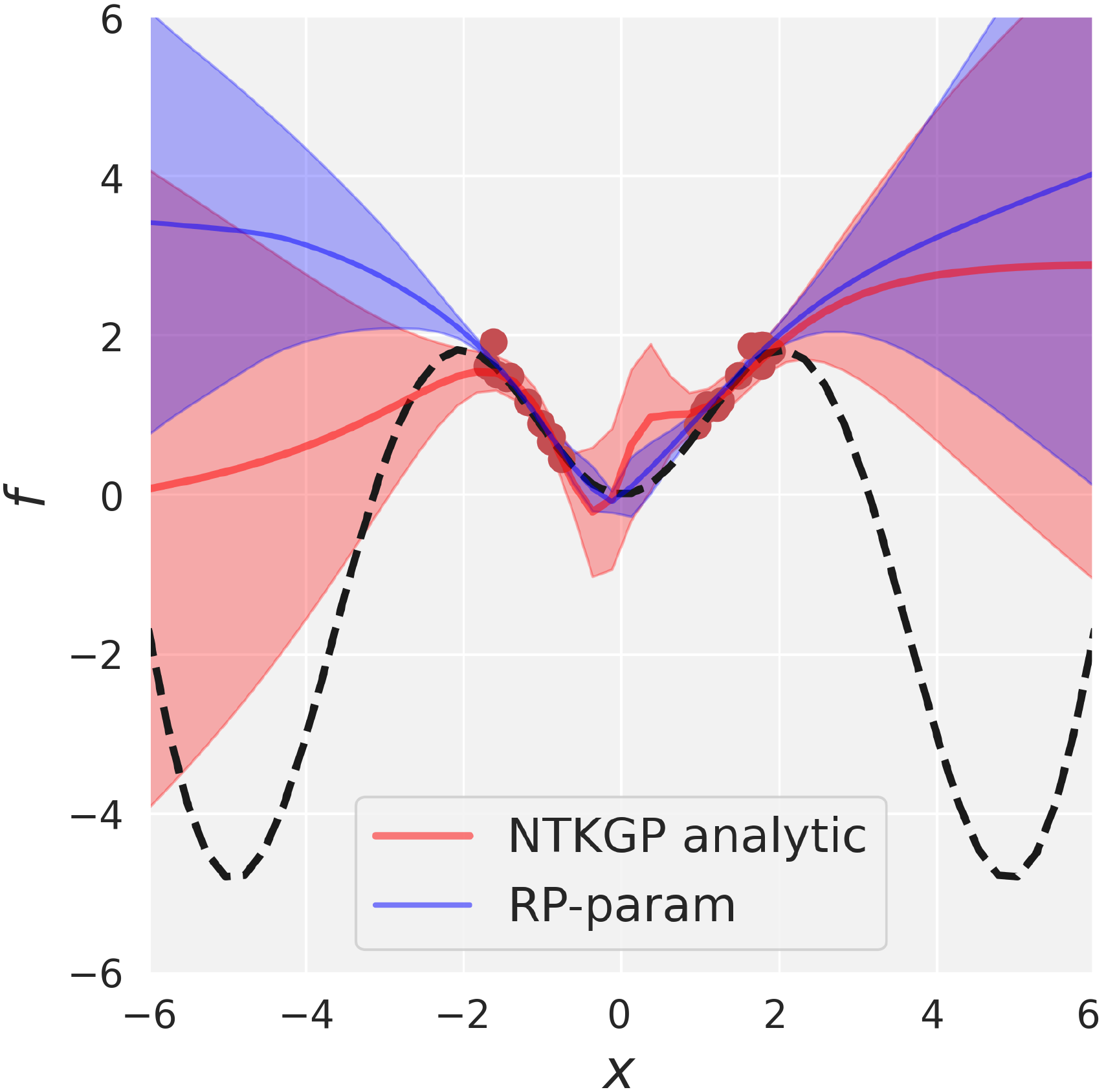}
  \caption{RP-param}
  \label{fig:xsinxRP}
\end{subfigure}
\begin{subfigure}{.15\textwidth}
  \centering
  \includegraphics[width=\linewidth]{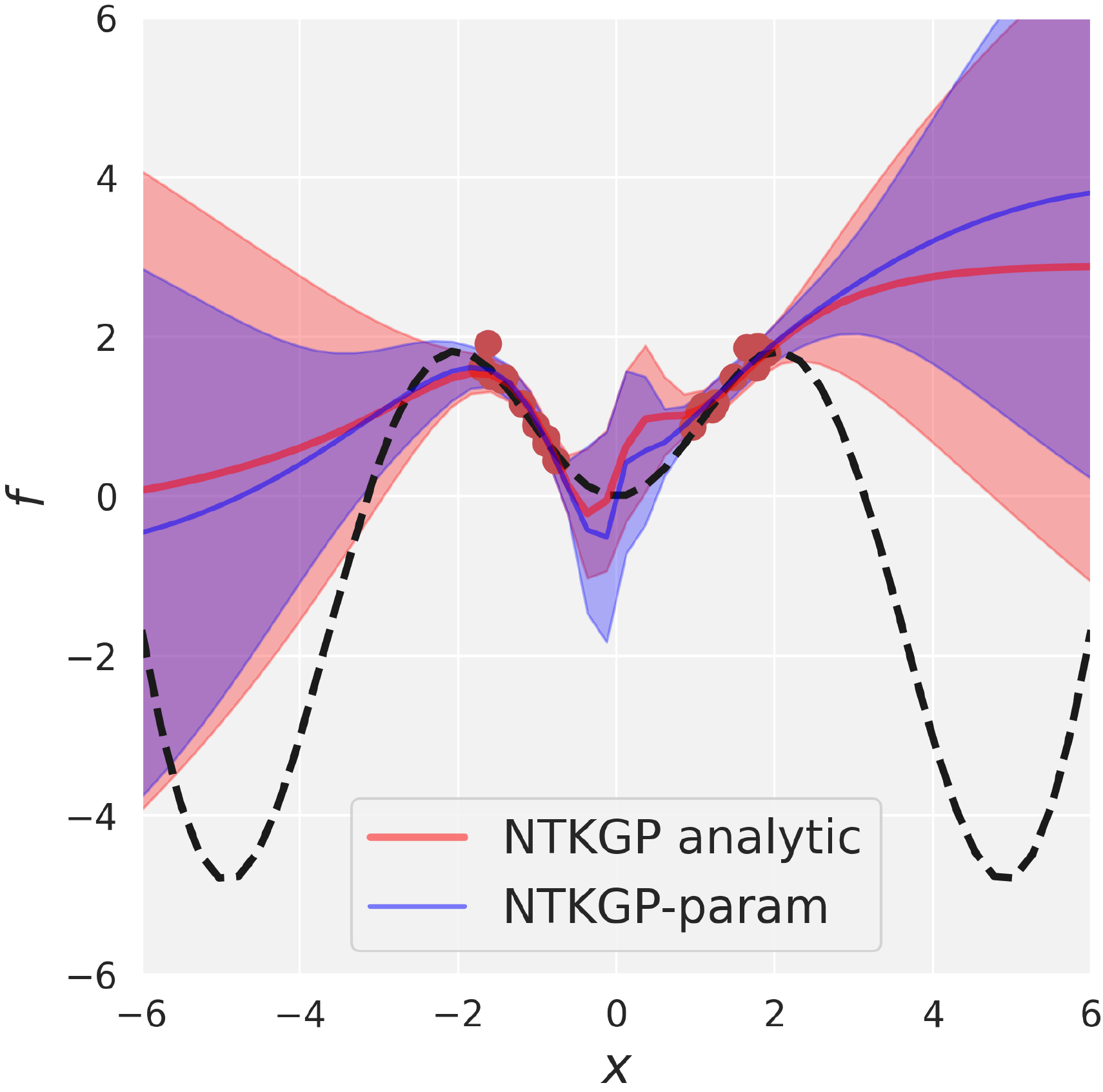}
  \caption{NTKGP-param}
  \label{fig:xsinxNTKGP}
\end{subfigure}
\begin{subfigure}{.15\textwidth}
  \centering
  \includegraphics[width=\linewidth]{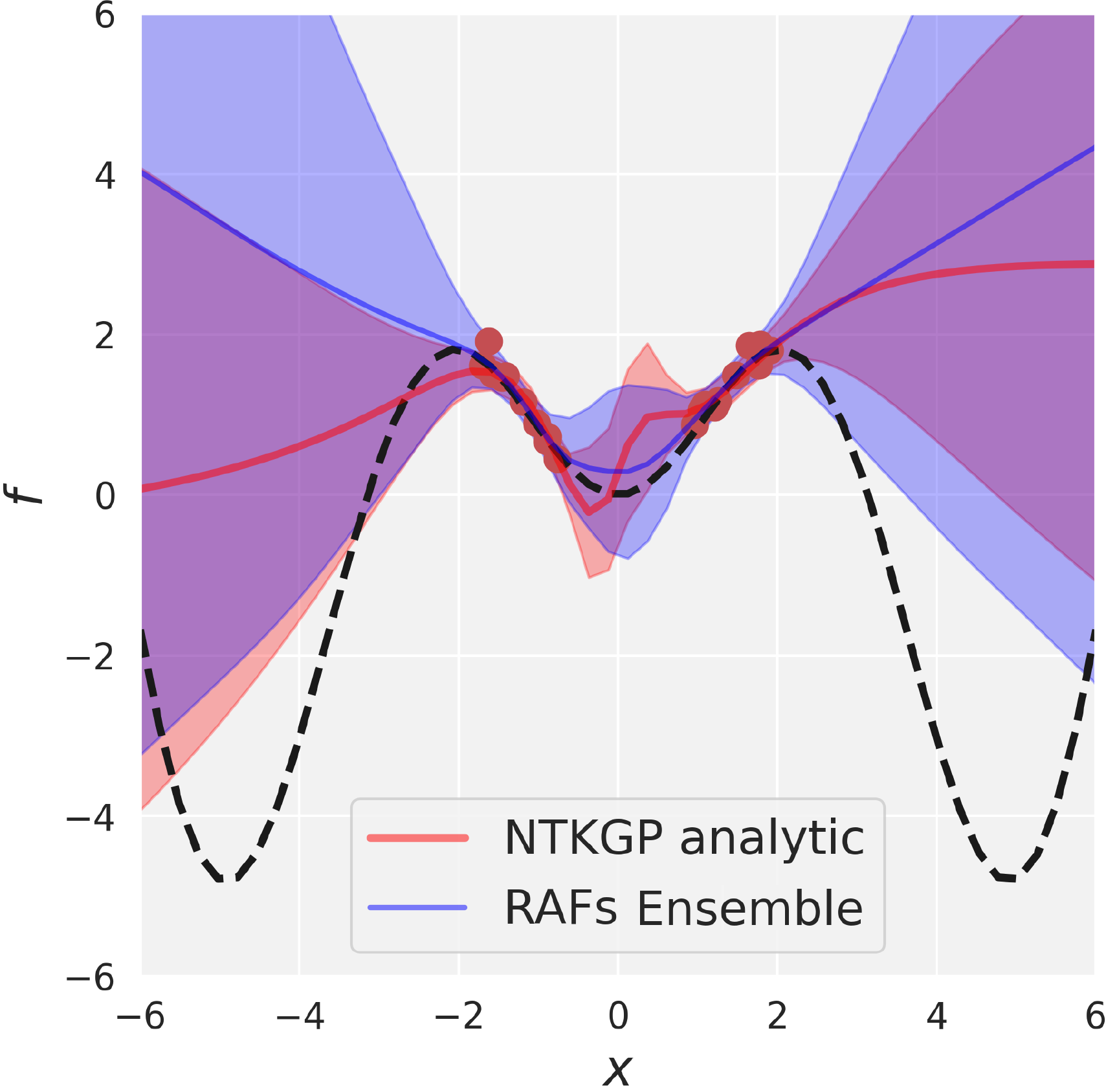}
  \caption{RAFs}
  \label{fig:xsinxRAFs}
\end{subfigure}
\caption{Uncertainty quantification of different methods on He et al. dataset. Gaussian process with neural tangent kernel (NTKGP analytic) is included as a reference.}
\label{fig:xsinx}
\end{figure}

\begin{table*}[t]
\centering
{\small
\begin{tabular}{l c c c c c c}
\hline
 &  & & NLL & & & \\
 \hline
 & DE & HDE & AE & NTKGP-p. & RP-p. & RAFs \\
 \hline
 He et al.  1D   & $>$100 $\pm$ 0.18 & 71.31 $\pm$ 0.51 & 38.75 $\pm$ 0.12 & 4.48 $\pm$ 0.18 & 13.05 $\pm$ 0.43 & \textbf{2.21 $\pm$ 0.18}\\

 Forrester et al. 1D   & $>$100 $\pm$ 0.53 & $>$100 $\pm$ 0.51 & 50.82 $\pm$ 0.52 & $>$100 $\pm$ 0.50 & 13.7 $\pm$ 0.58 & \textbf{0.64 $\pm$ 0.74}\\

 Schaffer N.4 2D   & 0.29 $\pm$ 0.01 & -0.71 $\pm$  0.01 & 2.15 $\pm$ 0.01 & -0.55 $\pm$ 0.01 &  -0.36 $\pm$ 0.01 & \textbf{-0.79 $\pm$ 0.01} \\
 
 Double pendulum 2D   & 2.95 $\pm$ 0.05  & 2.18 $\pm$ 0.84 & -0.36 $\pm$ 0.05 & \textbf{-0.58 $\pm$ 0.05} &  \textbf{-0.47 $\pm$ 0.05} & \textbf{-0.49 $\pm$ 0.04} \\
 
 Rastrigin 3D & 29.24 $\pm$ 1.30  & \textbf{3.09 $\pm$  1.15} & 35.94 $\pm$ 0.74 & 28.38 $\pm$ 0.64 &  \textbf{4.35 $\pm$ 1.24} & \textbf{3.44 $\pm$ 1.05} \\
 
 Ishigami 3D   & 6.01 $\pm$ 0.08  & $>$100 $\pm$  0.08 & 8.73 $\pm$ 0.08 & 1.53 $\pm$ 0.08 &  \textbf{-0.01 $\pm$ 0.08} & \textbf{0.06  $\pm$ 0.07} \\
 
 Environmental 4D & 64.72 $\pm$ 0.23  & 7.84 $\pm$  0.13 & 1.65 $\pm$ 0.20 & 4.5 $\pm$ 0.27 &  3.94 $\pm$ 0.21 & \textbf{0.81 $\pm$ 0.17} \\
 
 Griewank 4D   & 28.29 $\pm$ 2.43  & \textbf{5.50 $\pm$  1.62} & \textbf{4.64 $\pm$ 3.06} & 10.21 $\pm$ 2.37 &  \textbf{4.29 $\pm$ 2.93} & \textbf{4.79 $\pm$ 2.40} \\
 
 Roos \& Arnold 5D   & -2.02 $\pm$ 0.01  & \textbf{-2.21 $\pm$  0.00} & -1.89 $\pm$ 0.01 & -1.71 $\pm$ 0.01 &  -1.70 $\pm$ 0.01 & -2.1 $\pm$ 0.01 \\
 
 Friedman 5D  & 96.94 $\pm$ 0.41  & $>$100 $\pm$  0.51 & 15.04 $\pm$ 0.50 & 41.69 $\pm$ 0.39 &  4.22 $\pm$ 0.44 & \textbf{1.78 $\pm$ 0.39} \\
 
 Planar arm torque 6D   & 9.58 $\pm$ 0.07  & 4.11 $\pm$  0.08 & 3.07 $\pm$ 0.05 & \textbf{-0.32 $\pm$ 0.08} &  -0.05 $\pm$ 0.07 & -0.16 $\pm$ 0.06 \\
 
 Sum of powers 6D & $>$100 $\pm$ 0.41  & $>$100 $\pm$  0.62 & 55.03 $\pm$ 0.43 & $>$100 $\pm$ 0.41 &  41.59 $\pm$ 0.40 & \textbf{35.22 $\pm$ 0.35} \\
 
 Ackley 7D  & 7.11 $\pm$ 0.23  & \textbf{1.38 $\pm$  0.16} & 2.50 $\pm$ 0.36 & 3.11 $\pm$ 0.27 &  2.09 $\pm$ 0.26 & \textbf{1.16 $\pm$ 0.08} \\
 
 Piston simulation 7D  & \textbf{-2.19 $\pm$ 0.00}  & 14.06 $\pm$  0.00 & 3.50 $\pm$ 2.40 & 2.87 $\pm$ 2.93 &  2.67 $\pm$ 0.42 & 3.63 $\pm$ 0.57 \\
 
 Robot arm 8D   & 10.71 $\pm$ 0.03  & 6.87 $\pm$  0.01 & 7.11 $\pm$ 0.01 & \textbf{0.27 $\pm$ 0.03} &  0.80 $\pm$ 0.06 & \textbf{0.25 $\pm$ 0.02} \\
 
 Borehole 8D    & $>$100 $\pm$ 1.01  & $>$100 $\pm$  1.01 & \textbf{4.89 $\pm$ 1.87} & \textbf{5.48 $\pm$ 3.54} &  \textbf{4.06 $\pm$ 1.20} & \textbf{4.36 $\pm$ 1.26} \\
 
 Styblinski-Tang 9D  & $>$100 $\pm$ 3.05  & $>$100 $\pm$  0.00 & 40.80 $\pm$ 5.33 & $>$100 $\pm$ 3.03 &  \textbf{15.82 $\pm$ 6.31} & 2\textbf{5.23 $\pm$ 4.12} \\
 
 PUMA560 9D    & 6.59 $\pm$ 0.15  & \textbf{1.62 $\pm$  0.14} & 4.24 $\pm$ 0.14 & 5.93 $\pm$ 0.08 &  6.40 $\pm$ 0.14 & 2.14 $\pm$ 0.13 \\
 
 Adapted Welch 10D  & $>$100 $\pm$ 0.81  & $>$100 $\pm$  0.75 & $>$100 $\pm$ 0.55 & $>$100 $\pm$ 0.75 &  $>$100 $\pm$ 0.57 & \textbf{78.53 $\pm$ 0.67} \\
 
 Wing weight 10D  & $>$100 $\pm$ 0.00  & 27.31 $\pm$  4.37 & \textbf{5.46 $\pm$ 4.36} & 67.30 $\pm$ 0.53 &  \textbf{5.54 $\pm$ 4.15} & \textbf{5.39 $\pm$ 1.69} \\
 \hline
 Boston housing & 74.54 $\pm$ 1.06 & $>$100 $\pm$ 1.04 & 71.53 $\pm$ 1.06 & 70.82 $\pm$ 1.06 & $>$100 $\pm$ 1.10 & \textbf{ 40.67 $\pm$ 1.00} \\
 
 Abalone & $>$100 $\pm$ 0.10 & $>$100 $\pm$ 0.10 & 47.67 $\pm$ 0.10 & $>$100 $\pm$ 0.10 & \textbf{28.37 $\pm$ 0.10} & 28.90 $\pm$ 0.10 \\
 
 Naval propulsion & \textbf{-2.27 $\pm$ 0.00} & $>$100 $\pm$ 0.00 & 3.92 $\pm$ 0.10 & 2.28 $\pm$ 1.51 & 2.16 $\pm$ 0.16 & 1.91 $\pm$ 0.07 \\
 
 Forest fire & 15.71  $\pm$ 0.05 & \textbf{3.14 $\pm$ 0.02} & \textbf{2.66 $\pm$ 0.69} &   \textbf{3.10 $\pm$ 1.11} & 4.68 $\pm$ 0.14 & \textbf{2.15 $\pm$ 0.28} \\
 
 Parkinson's & \textbf{26.74  $\pm$ 0.02} & $>$100 $\pm$ 0.10 & $>$100 $\pm$ 0.16 &   $>$100 $\pm$ 0.03 & $>$100 $\pm$ 0.16 & 45.69 $\pm$ 0.16 \\
\hline
\end{tabular}}
\caption{Performance of methods on all datasets w.r.t. NLL, including 95\% confidence intervals. The best scores are in bold. \label{tab:nll}}
\end{table*}

\subsection{Performance Results}
\subsubsection{Qualitative Comparison.}
Figure \ref{fig:xsinx} shows the performance of different methods compared to a Gaussian process with a neural tangent kernel (NTKGP analytic) as a reference, on a 1D toy dataset generated from $\boldsymbol{y} = \boldsymbol{x}\text{sin}(\boldsymbol{x}) + \boldsymbol{\epsilon}$ (dashed line). 
%
The plots demonstrate that DE, HDE, and AE provide narrow uncertainty bounds in areas where no data has been observed by the model, which translates to high confidence in OOD data. On the contrary, NTKGP-param, RP-param, and RAFs Ensemble bound their uncertainty estimates with wider intervals in areas with no data, accounting for adequate quantification of epistemic uncertainty, while also indicating robustness to OOD data. Among these methods, RAFs Ensemble provides the widest uncertainty which is reasonable considering the amount of data that is available to the methods over each area. Moreover, this observation is quantitatively validated as RAFs Ensemble achieves the lowest NLL compared to the other methods (see Table \ref{tab:nll}). 

\subsubsection{Overall Performance.}
We evaluate the overall performance of all methods in terms of both NLL and RMSE. The outcomes of comparing RAFs Ensemble with five baseline methods on twenty synthetic and five real-world datasets are outlined in Table~\ref{tab:nll} and Table~\ref{tab:rmse}. The results illustrate that our approach outperforms the competitors in most scenarios. Furthermore, Table~\ref{tab:rank} summarizes the obtained results in terms of ranking, in which the methods are ranked based on their performance for a particular dataset. 
The left integer corresponds to NLL, while the right one points to RMSE, and the bold values indicate the best-performing method.


\subsubsection{Discussion.}
The obtained results in this section illustrate that DE has good uncertainty estimates with respect to NLL for the real-world datasets, and it takes the first place for Naval propulsion and Parkinson's datasets. For the rest of the data categories, when compared to the other methods, DE fails to provide strong performance, usually scoring a very low NLL rank. Therefore, this indicates that Deep Ensemble might have difficulty quantifying epistemic uncertainty in general as displayed by the experiments in this paper, but seemingly manages to capture aleatoric uncertainty well.

Unlike DE, the HDE provides outwardly reliable estimates for datasets with many local minima, despite its unimpressive overall results when compared to the other methods. However, both DE and HDE can produce uncertainty bounds that are unreasonably narrow in areas with unobserved data, as shown in Figure \ref{fig:xsinx} and noted by \cite{Heiss2021NOMUNO}.

\begin{table*}[t]
\centering
{\small
\begin{tabular}{l c c c c c c}
\hline
 &  & & RMSE & & & \\
 \hline
 & DE & HDE & AE & NTKGP-p. & RP-p. & RAFs\\
 \hline
 He et al.  1D   & 3.71 $\pm$ 0.18 & 5.70 $\pm$ 0.51 & \textbf{3.15 $\pm$ 0.12} & 3.64 $\pm$ 0.18 & 5.24 $\pm$ 0.43 & 3.80 $\pm$ 0.18\\

 Forrester et al. 1D   & 5.00 $\pm$ 0.53 & 4.12 $\pm$ 0.51 & 4.09 $\pm$ 0.52 & 6.05 $\pm$ 0.50 & 5.70 $\pm$ 0.58 & \textbf{2.8 $\pm$ 0.74}\\

 Schaffer N.4 2D   & \textbf{0.23 $\pm$ 0.01} & 0.34 $\pm$  0.01 & 0.30 $\pm$ 0.01 & \textbf{0.24 $\pm$ 0.01} &  0.31 $\pm$ 0.01 & 0.27 $\pm$ 0.01 \\
 
 Double pendulum 2D   & \textbf{0.46 $\pm$ 0.05}  & 2.22 $\pm$ 0.84 & 0.71 $\pm$ 0.05 & \textbf{0.51 $\pm$ 0.05} &  0.74 $\pm$ 0.05 & \textbf{0.58 $\pm$ 0.04} \\
 
 Rastrigin 3D & 18.41 $\pm$ 1.30  & \textbf{10.96 $\pm$  1.15} & 25.58 $\pm$ 0.74 & 18.10 $\pm$ 0.64 &  \textbf{12.87 $\pm$ 1.24} & \textbf{14.85 $\pm$ 1.05} \\
 
 Ishigami 3D   & \textbf{0.69 $\pm$ 0.08}  & 1.05 $\pm$  0.08 & 0.88 $\pm$ 0.08 & \textbf{0.69 $\pm$ 0.08} &  \textbf{0.58 $\pm$ 0.08} & \textbf{0.57  $\pm$ 0.07} \\
 
 Environmental 4D & \textbf{2.04 $\pm$ 0.23}  & 2.51 $\pm$  0.13 & \textbf{1.83 $\pm$ 0.20} & \textbf{2.34 $\pm$ 0.27} &  \textbf{2.03 $\pm$ 0.21} & \textbf{1.68 $\pm$ 0.17} \\
 
 Griewank 4D   & 83.97 $\pm$ 2.43  & \textbf{45.68 $\pm$  1.62} & \textbf{42.12 $\pm$ 3.06} & 78.47 $\pm$ 2.37 &  \textbf{38.62 $\pm$ 2.93} & 78.79 $\pm$ 2.40 \\
 
 Roos \& Arnold 5D   & 0.07 $\pm$ 0.01  & \textbf{0.01 $\pm$  0.00} & 0.07 $\pm$ 0.01 & 0.09 $\pm$ 0.01 &  0.08 $\pm$ 0.01 & 0.08 $\pm$ 0.01 \\
 
 Friedman 5D  & \textbf{3.17 $\pm$ 0.41}  & \textbf{3.63 $\pm$  0.51} & \textbf{2.95 $\pm$ 0.50} & \textbf{3.39 $\pm$ 0.39} &  \textbf{2.74 $\pm$ 0.44} & \textbf{3.1 $\pm$ 0.39} \\
 
 Planar arm torque 6D   & \textbf{0.65 $\pm$ 0.07}  & \textbf{0.62 $\pm$  0.08} & \textbf{0.71 $\pm$ 0.05} & \textbf{0.71 $\pm$ 0.08} &  1.08 $\pm$ 0.07 & \textbf{0.74 $\pm$ 0.06} \\
 
 Sum of powers 6D & \textbf{22.81 $\pm$ 0.41}  & \textbf{21.19 $\pm$  0.62} & \textbf{21.87 $\pm$ 0.43} & \textbf{22.79 $\pm$ 0.41} &  \textbf{22.22 $\pm$ 0.40} & \textbf{22.24 $\pm$ 0.35} \\
 
 Ackley 7D  & 8.92 $\pm$ 0.23  & 2.43 $\pm$  0.16 & 7.28 $\pm$ 0.36 & 8.58 $\pm$ 0.27 &  4.03 $\pm$ 0.26 & \textbf{1.33 $\pm$ 0.08} \\
 
 Piston simulation 7D  & \textbf{0.02 $\pm$ 0.00}  & 0.04 $\pm$  0.00 & 29.1 $\pm$ 2.40 & $>$100 $\pm$ 2.93 &  5.78 $\pm$ 0.42 & 7.40 $\pm$ 0.57 \\
 
 Robot arm 8D   & 0.92 $\pm$ 0.03  & \textbf{0.80 $\pm$ 0.01}  & 0.88 $\pm$ 0.01 & 0.93 $\pm$ 0.03 &  1.09 $\pm$ 0.06 & \textbf{0.83 $\pm$ 0.02} \\
 
 Borehole 8D    & \textbf{32.11 $\pm$ 1.01}  & \textbf{32.12 $\pm$  1.01} & 48.75 $\pm$ 1.87 & $>$100 $\pm$ 3.54 &  38.60 $\pm$ 1.20 & 41.35 $\pm$ 1.26 \\
 
 Styblinski-Tang 9D  & $>$100 $\pm$ 3.05  & $>$100 $\pm$  0.00 & \textbf{$>$100 $\pm$ 5.33} & $>$100 $\pm$ 3.03 &  \textbf{$>$100 $\pm$ 6.31} & $>$100 $\pm$ 4.12 \\
 
 PUMA560 9D    & 3.93 $\pm$ 0.15  & \textbf{3.23 $\pm$  0.14} & \textbf{3.40 $\pm$ 0.14} & 3.93 $\pm$ 0.08 &  \textbf{3.24 $\pm$ 0.14} & \textbf{3.4 $\pm$ 0.13} \\
 
 Adapted Welch 10D  & \textbf{99.51 $\pm$ 0.81}  & \textbf{99.4 $\pm$  0.75} & $>$100 $\pm$ 0.55 & \textbf{99.79 $\pm$ 0.75} &  $>$100 $\pm$ 0.57 & \textbf{100.00 $\pm$ 0.67} \\
 
 Wing weight 10D  & $>$100 $\pm$ 0.00  & \textbf{58.16 $\pm$  4.37} & \textbf{63.1 $\pm$ 4.36} & $>$100 $\pm$ 0.53 &  \textbf{63.35 $\pm$ 4.15} & $>$100 $\pm$ 1.69 \\
 \hline
 Boston housing & \textbf{11.28 $\pm$ 1.06} & \textbf{11.36 $\pm$ 1.04} & \textbf{11.42 $\pm$ 1.06} & \textbf{11.28 $\pm$ 1.06} & \textbf{11.56 $\pm$ 1.10} & \textbf{11.31 $\pm$ 1.00} \\
 
 Abalone & \textbf{2.06 $\pm$ 0.10} & \textbf{2.09 $\pm$ 0.10} & \textbf{2.08 $\pm$ 0.10} & \textbf{2.05 $\pm$ 0.10} & \textbf{2.09 $\pm$ 0.10} & \textbf{2.08 $\pm$ 0.10} \\
 
 Naval propulsion &  \textbf{0.02 $\pm$ 0.00} & \textbf{0.02 $\pm$ 0.00} & 38.86 $\pm$ 0.60 & 62.61 $\pm$ 1.51 & 9.40 $\pm$ 0.16 & 3.45 $\pm$ 0.08 \\
 
 Forest fire & 1.97  $\pm$ 0.05 & \textbf{1.87 $\pm$ 0.02} & 6.43 $\pm$ 0.69 &   10.43 $\pm$ 1.11 & 2.32 $\pm$ 0.14 & 3.32 $\pm$ 0.28 \\
 
 Parkinson's & 12.17  $\pm$ 0.02 & 12.40 $\pm$ 0.10 & 12.49 $\pm$ 0.16 &   \textbf{11.97 $\pm$ 0.03} & 12.60 $\pm$ 0.16 & 12.78 $\pm$ 0.16 \\
 
\hline
\end{tabular}}
\caption{Performance of methods on all datasets w.r.t. RMSE, including 95\% confidence intervals. The best scores are in bold. \label{tab:rmse}}
\end{table*}

\begin{table}[!ht]
\centering
\resizebox{\columnwidth}{!}{
\begin{tabular}{l c c c c c c}
    & \text{DE}        & \text{HDE}       & \text{AE}       & \text{NTKGP-p.}       & \text{RP-p.}    & \text{RAFs} \\
He et al. 1D & 6,2 & 5,3 & 4,1 & 2,2 & 3,3 & \textbf{1},2\\ 

Forrester et al. 1D & 4,2 & 5,2 & 3,2 & 6,2  & 2,2 &  \textbf{1},1 \\

Schaffer N.4 2D & 5,1 & 2,4 & 6,3 & 3,1 & 4,3 & \textbf{1},2 \\

Double pendulum 2D & 3,1 & 3,3 & 2,2 & \textbf{1},1 & \textbf{1},2 & \textbf{1},1 \\

Rastrigin 3D & 2,2 & \textbf{1},1 & 3,3 & 2,2 & \textbf{1},1 & \textbf{1},1 \\

Ishigami 3D & 3,1 & 5,3 & 4,2 & 2,1 & \textbf{1},1 & \textbf{1},1 \\

Environmental 4D & 6,1 & 5,2 & 2,1 & 4,1 & 3,1 & \textbf{1},1 \\

Griewank 4D & 3,3 & \textbf{1},1 & \textbf{1},1 & 2,2 & \textbf{1},1 & \textbf{1},2 \\

Roos \& Arnold 5D & 3,1 & \textbf{1},1 & 4,3 & 5,1 & 5,1 & 2,2 \\

Friedman 5D & 5,1 & 6,1 & 3,1 & 4,1 & 2,1 & \textbf{1},1 \\

Planar arm torque 6D & 5,1 & 4,1 & 3,1 & \textbf{1},1 & 2,2 & 2,1 \\

Sum of powers 6D & 5,1 & 6,1 & 3,1 & 4,1 & 2,1 & \textbf{1},1 \\

Ackley 7D & 4,5 & \textbf{1},2 & 2,4 & 3,5 & 2,3 & \textbf{1},1 \\

Piston simulation 7D & \textbf{1},1 & 3,2 & 2,5 & 2,6 & 2,3 & 2,4 \\

Robot arm 8D & 5,3 & 3,1 & 4,2 & \textbf{1},3 & 2,4 & \textbf{1},1 \\

Borehole 8D & 2,1 & 3,1 & \textbf{1},4 & \textbf{1},5 & \textbf{1},2 & \textbf{1},3 \\

Styblinski-Tang 9D & 3,2 & 5,5 & 2,1 & 4,3 & \textbf{1},1 & \textbf{1},4 \\

PUMA560 9D & 5,2 & \textbf{1},1 & 3,1 & 4,2 & 4,1 & 2,1 \\

Adapted Welch 10D & 6,1 & 2,1 & 4,3 & 5,1 & 3,2 & \textbf{1},1 \\

Wing weight 10D & 4,4 & 2,1 & \textbf{1},1 & 3,3 & \textbf{1},1 & \textbf{1},2 \\
\hline
Boston housing & 3,1 & 5,1 & 2,1 & 2,1 & 5,1 & \textbf{1},1 \\

Abalone & 5,1 & 6,1 & 3,1 & 4,1 & \textbf{1},1 & 2,1 \\

Naval propulsion plant & \textbf{1},1 & 4,1 & 3,4 & 2,5 & 2,3 & 2,2 \\

Forest fire & 3,2 & \textbf{1},1 & 1,5 & 1,6 & 2,3 & 1,4 \\

Parkinson's & \textbf{1},2 & 6,3 & 4,3 & 5,1 & 3,3 & 2,3
\end{tabular}}
\caption{Rank of the methods corresponding to NLL (left) and RMSE (right). The best overall score is in bold (ties are possible in case of an overlap in confidence intervals). \label{tab:rank}}
\end{table} 

Nonetheless, AE demonstrates good performance in the dataset categories that exhibit higher epistemic uncertainty such as the physical models. This is due to the fact that AE is designed for capturing model uncertainty, while aleatoric uncertainty is assumed to be constant. Accordingly, AE achieves inferior performance on the real-world datasets, as those generally have more data uncertainty appropriated.

On the other hand, NTKGP-param achieves its finest performance for datasets in the physical model category, which is normally associated with substantial model uncertainty. A credible rationale to explain this insight is the fact that NTKGP-param tends to be more conservative than Deep Ensemble. However, it is generally unclear in which situations this is beneficial since the ensemble members of NTKGP-param will always be misspecified in practice according to \cite{He2020BayesianDE}.

Furthermore, RP-param manages to rank comparatively high for real-world datasets as well as trigonometric data, that contain vast amounts of aleatoric and epistemic uncertainty, respectively, indicating that it does not quantify either type of uncertainty better than the other. This observation serves as a demonstration that RP-param generalizes well for different types of datasets that exhibit broad characteristics. However, this technique fails to deliver low NLL scores on some occasions, which might be attributed to the fact that RP-param is based on bootstrapping. While bootstrapping can be a successful strategy for inducing diversity, it can sometimes harm the performance when the base-learners have multiple local optima, as is a common case  with NNs \cite{Lakshminarayanan2017SimpleAS}.

Nevertheless, RAFs Ensemble outperforms RP-param, and every other method in the comparisons, for 13 out of 25 datasets. In terms of NLL, our approach does not rank below the second place for any data, which is consistent with the strong results from Table \ref{tab:nll}. Meanwhile, the RMSE scores of this method are altogether satisfactory, although not as prominent compared to the NLL scores. In agreement with the overall outstanding results, RAFs Ensemble holds the highest NLL rank for all data from \textit{MLM} and \textit{T} categories, which can be contemplated as a concluding statement regarding its capabilities to estimate epistemic uncertainty and challenging multimodality. Among all categories, the real-world datasets are least favored by RAFs Ensemble, primarily due to their high level of aleatoric uncertainty. This indicates that RAFs Ensemble captures model uncertainty slightly better than aleatoric uncertainty. Nonetheless, the empirical superiority of this technique is due to the exhaustively exploited added source of randomness via random activation functions, combined with method simplicity and Bayesian behavior, resulted from the anchored loss (Equation \ref{anchor}). This successful combination leads to greatly improved diversity among ensemble members, which can be further confirmed by a direct comparison between RAFs Ensemble and AE. 
Note that even though RAFs Ensemble does not provide as prominent results with respect to RMSE in the higher dimensional datasets as it does in datasets of lower dimensions, it still achieves better or on par results compared to the state-of-the-art methods. In addition, RAFs Ensemble can be deployed in both complex and straightforward settings. On a related note, while  DE struggles when dealing with high multimodality and RP-param underperforms when the dataset has interaction effects (from ``others" category), RAFs excels in both such settings.

\subsubsection{Scalability to higher dimensions and larger networks.} 
To test the scalability of RAFs Ensemble, we compare it with the strongest baseline, RP-param, on two additional real-world datasets, i.e., a 65-dimensional data with around 20k samples and a 40-dimensional data with almost 40k samples. The former is the superconductivity dataset, where the goal is to predict the critical temperature of superconductors \cite{Hamidieh2018ADS}. The latter summarizes features about articles, where the target is the number of shares in social networks \cite{Fernandes2015API}. Both methods utilize the same neural architecture for their base-learners,
that is two hidden layers of 128 hidden neurons each, which is more complex than the previous experiments. The conclusion of these experiments is conclusive in favor of our approach. 
RAFs Ensemble scores NLL of 5.49 and 25.89 on the first and second dataset, respectively, while RP-param scores NLL of over 100 on both datasets.

\begin{figure}[t!]
\centering
\begin{subfigure}{.30\textwidth}
  \centering
  \includegraphics[width=1\linewidth]{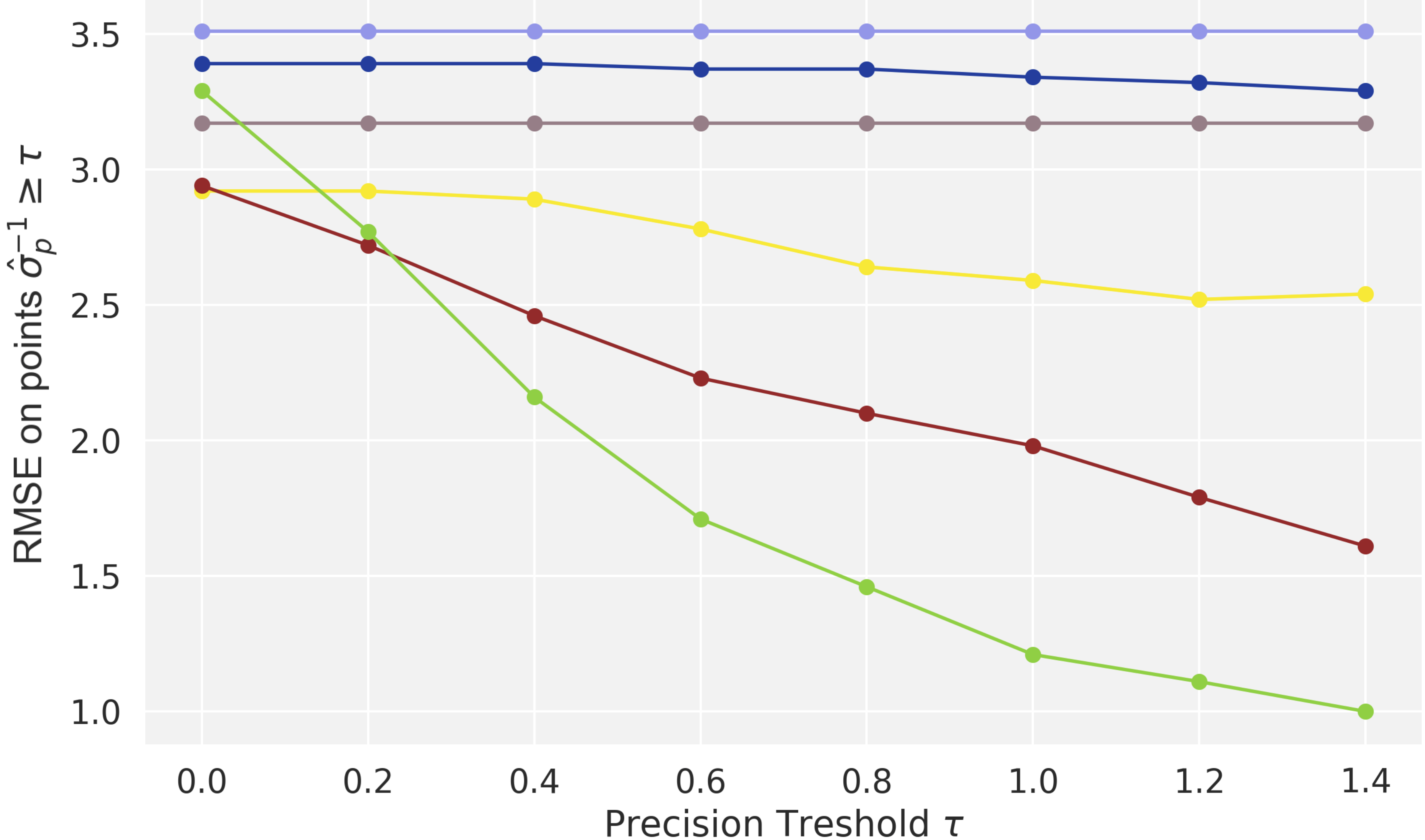}
  \caption{Friedman (synthetic)}
  \label{fig:conferr_friedman}
\end{subfigure}%
\vspace{1mm}
\begin{subfigure}{.30\textwidth}
  \centering
  \includegraphics[width=1\linewidth]{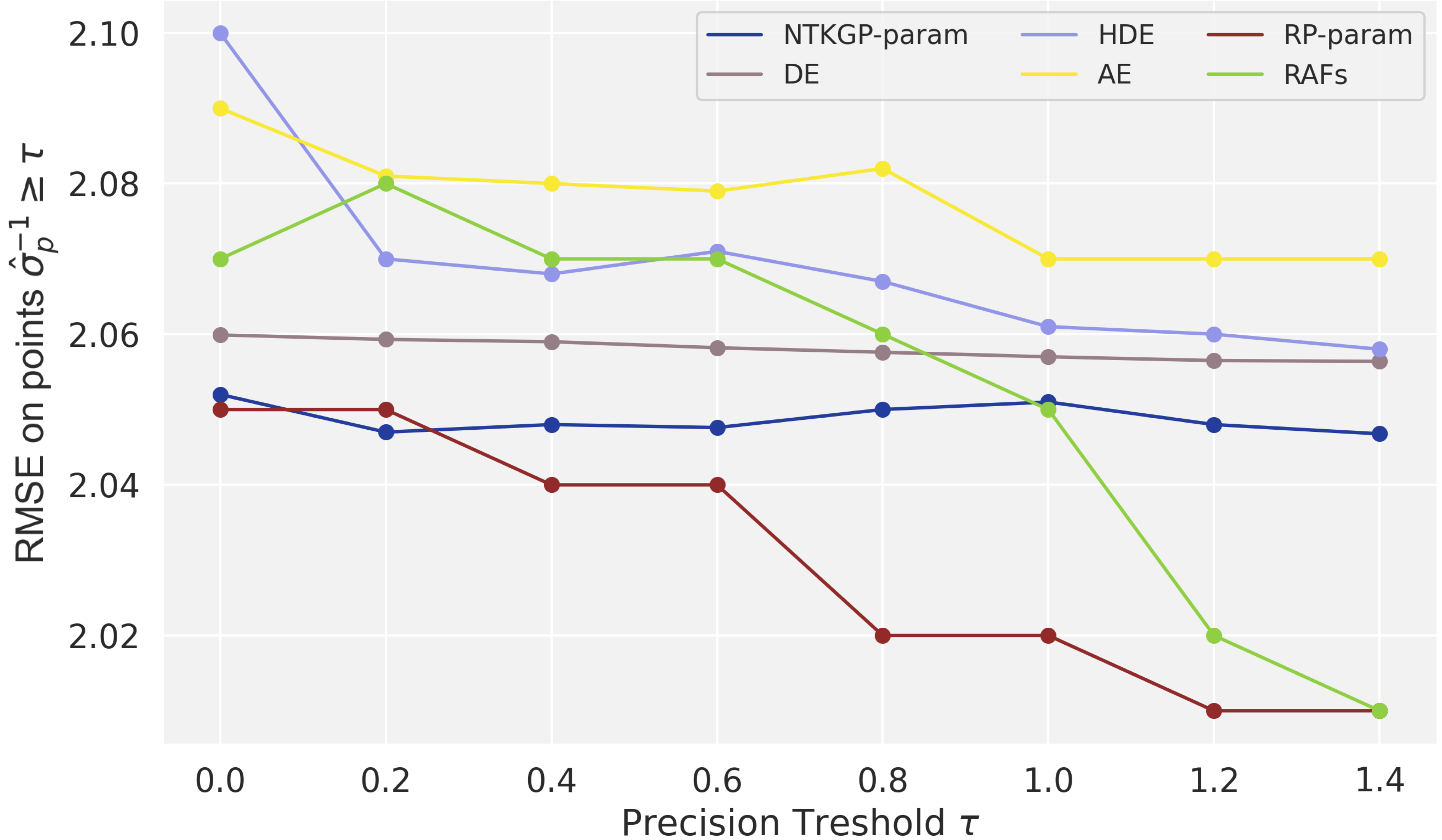}
  \caption{Abalone (real-world)}
  \label{fig:conferr_abalone}
\end{subfigure}
\caption{Confidence versus error of estimations.}
\label{fig:caler}
\end{figure}

\subsubsection{Confidence vs. Error.}
We further analyze the relation between the RMSE and the precision thresholds in order to examine the confidence of each method in the prediction task. Figure \ref{fig:caler} displays the confidence versus error plots for one synthetic and one real-world dataset, i.e., Friedman and Abalone (see the Technical Appendix for more detail). In this figure, for each precision threshold $\tau$, the RMSE is plotted for examples where the predicted precision $\boldsymbol{\sigma}_p^{-2}$ is larger than the threshold $\tau$, demonstrating confidence.
In general, reliable estimates are expected to have decreasing error when the confidence is increasing. For Friedman dataset, it is clear that RAFs Ensemble delivers well-calibrated estimates, which is especially in contrast with  DE, NTKGP-param, and HDE (Figure \ref{fig:conferr_friedman}). However, for the Abalone data, RP-param demonstrates the most reliable behavior, although RAFs Ensemble meets its performance at the last precision threshold (Figure \ref{fig:conferr_abalone}). Overall, our approach sustains lower error over most precision thresholds compared to the majority of the other methods, and this contrast in performance is emphasized as the predictions get more confident.

\subsubsection{Ablation.}
\begin{figure}[t!]
\begin{subfigure}{.2\textwidth}
  \centering
  \includegraphics[width=\linewidth]{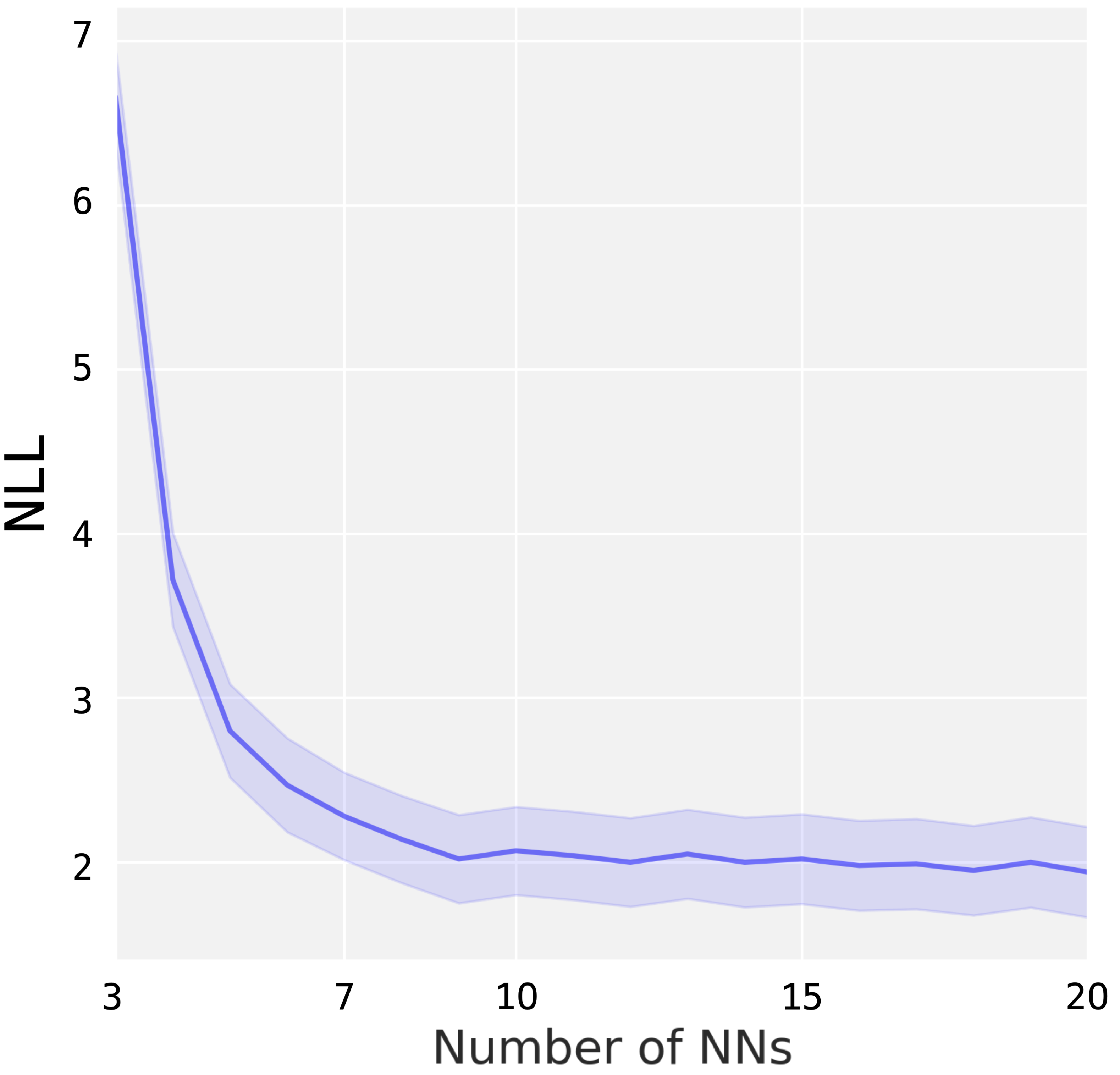}
  \caption{PUMA560}
  \label{fig:RAFsPUMA}
\end{subfigure}~
\begin{subfigure}{.22\textwidth}
  \centering
  \includegraphics[width=1.02\linewidth]{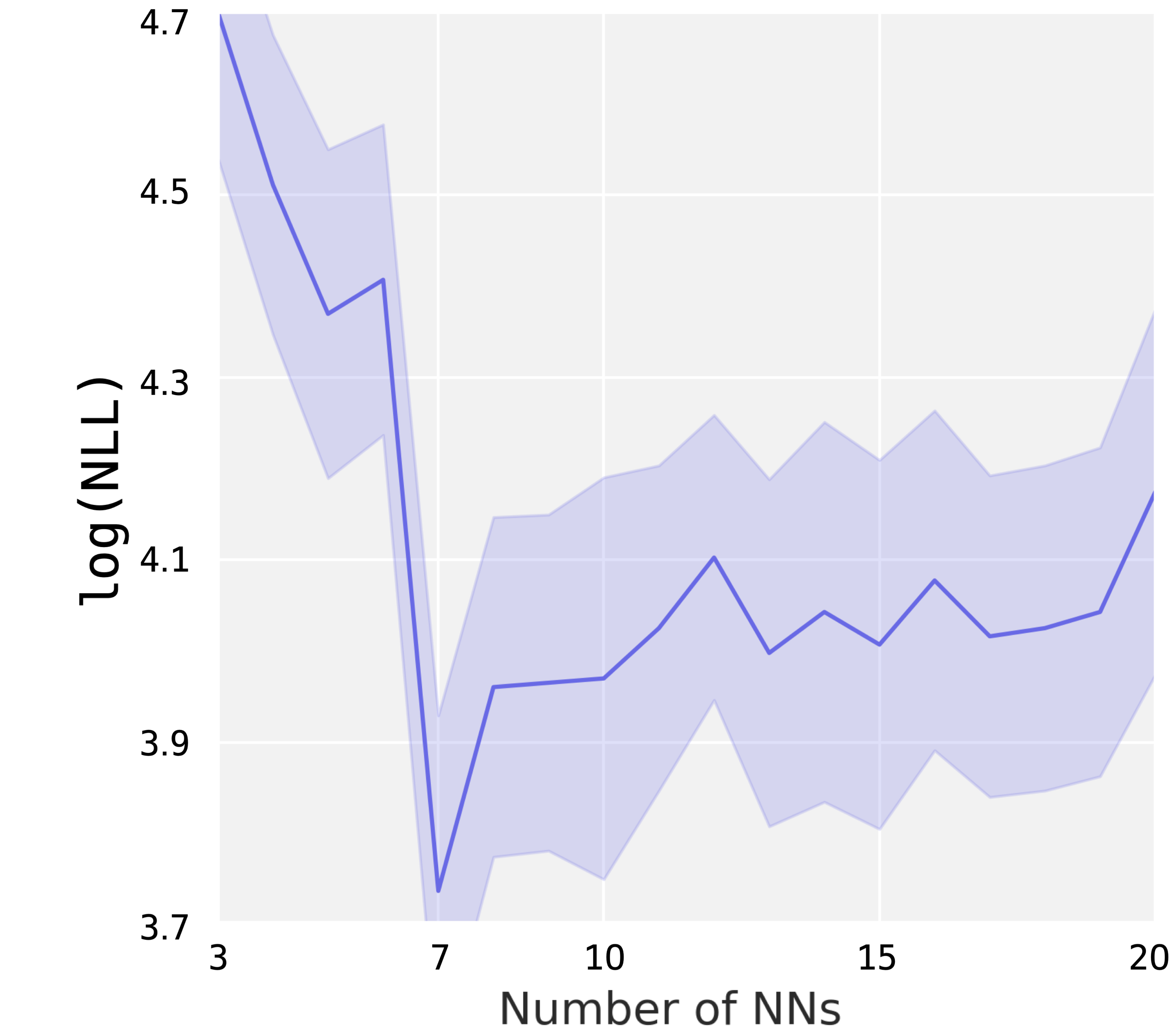}
  \caption{Abalone}
  \label{fig:RAFsAbalone}
\end{subfigure}
\caption{The effect of number of NNs in the ensemble in terms of NLL, including the 95\% confidence interval.}
\label{fig:exprafs}
\end{figure}

We study the effect of number of base-learners in the ensemble on the quality of UQ, which also measures the sensitivity of the results to the cardinality of the set of AFs $k$. We conduct an experiment on two different datasets, one synthetic (PUMA590) and one real-world (Abalone), where the results in terms of NLL are represented in Figure \ref{fig:exprafs}.
Note that Figure \ref{fig:RAFsAbalone} is shown in log-scale for better visibility. According to the theory, in the limit of infinite number of ensemble members, the ensemble error converges to zero \cite{Hansen1990NeuralNE}. However, practically speaking, five NNs in the ensemble provide optimal results regarding the trade-off between empirical performance and computational time \cite{Lakshminarayanan2017SimpleAS}, which is also the case in our experiments. This is further confirmed by the plot in Figure \ref{fig:RAFsPUMA}. In addition, for the PUMA590 dataset, it seems that RAFs Ensemble's performance is not impacted negatively by the number of NNs in the ensemble. Moreover, an interesting observation is the steep through for seven NNs (equal to $k$) in Figure \ref{fig:RAFsAbalone}, which is an indication that there might be a correlation between $k$ and the performance in some cases. A plausible reason for this is the fact that the additional source of randomness is utmostly exploited via a different activation function. 

\begin{wrapfigure}{r}{0.2\textwidth}
\includegraphics[width=0.2\textwidth]{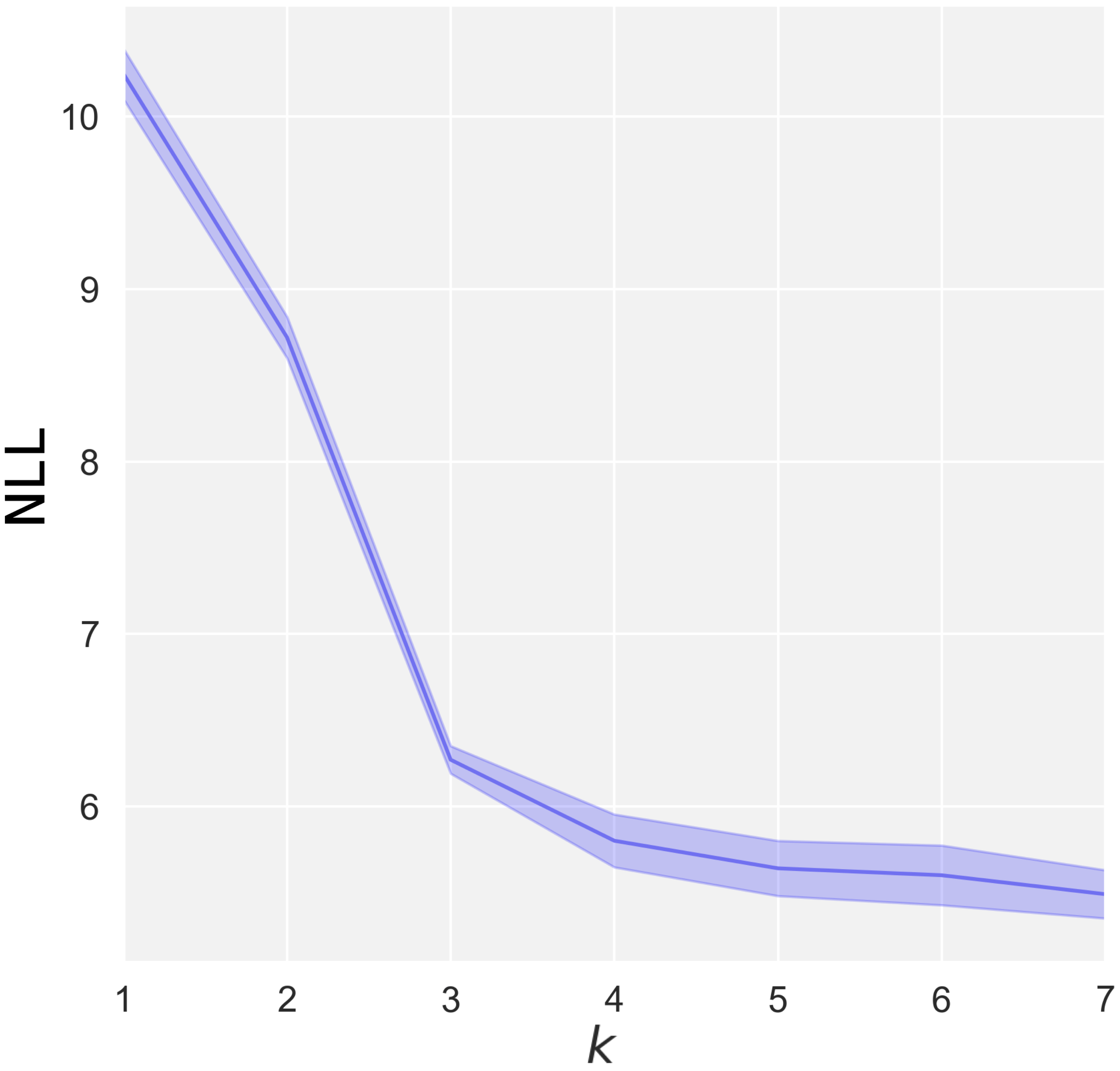}
\caption{The effect of cardinality $k$ on NLL for Superconductivity data.}
\label{fig:RAFsSupercond}
\end{wrapfigure} 

To further confirm the effectiveness of the random activation functions, we evaluate the performance of RAFs Ensemble (of five NNs) in terms of NLL w.r.t. different cardinalities $k$ of the set of AFs. The dataset used for this experiment is the superconductivity data. As the results in Figure \ref{fig:RAFsSupercond} clearly suggest, by increasing the cardinality $k$, NLL has a decreasing pattern, which shows that having more random AFs significantly improves the performance of the ensemble.

Moreover, we combine our approach with RP-param instead of AE to show that RAFs can be methodologically applied to any ensemble technique. We evaluate the performance of this combination on the Parkinson’s dataset, using the same network architecture for fair comparison.
The obtained results demonstrate that applying RAFs to RP-param leads to reducing the original NLL score of $>100$ to 48.66, which is in line with the results we get when comparing AE with RAFs Ensemble and is a further proof that the methodology indeed increases the performance.



\section{Conclusions}
We introduced a novel method, Random Activation Functions Ensemble, for a more robust uncertainty estimation in approaches based on neural networks, in which, each network in the ensemble is accomodated with a different (random) activation function to increase the diversity of the ensemble. The empirical study illustrates that our approach achieves excellent results in quantifying both epistemic and aleatoric uncertainty compared to five state-of-the-art ensemble uncertainty quantification methods on a series of regression tasks across 25 datasets, which proved there does not have to be a trade-off between simplicity and strong empirical performance. Furthermore, the properties of datasets such as dimensionality or complexity of modeling dynamics do not appear to affect RAFs Ensemble negatively, which also demonstrates robustness in out-of-distribution settings. 



\bibliography{aaai23.bib}
\newpage
\appendix
\section{Appendix}

\section{Synthetic datasets}
\label{synthdata}
The number of training data points and testing data points for each dataset is shown in Table \ref{numberpoints}. 
\subsection{Physical models}
Both the training data and testing data of every dataset in this dataset category are sampled from realistic ranges, so that all values are also possible in a real world setting. 

\subsubsection{Double pendulum 2D}
A double pendulum, is a dynamical system, which consists of a pendulum with another pendulum attached to its end \cite{Levien1993DoublePA}. For the purposes of this work, the length of both pendulum ropes $L_1$ and $L_2$ is kept constant $L_1 = L_2 = 1 $ and the response variable $y_i$ that is being modeled is the horizontal position of the lower pendulum mass given $\boldsymbol{\theta}_1$ and $\boldsymbol{\theta}_2$:

\begin{equation}
    \boldsymbol{y} = L_1 \sin(\boldsymbol{\theta}_{1}) + L_2 \sin(\boldsymbol{\theta}_{2}) + \epsilon
\end{equation} 

For generating the training dataset, the inputs $\boldsymbol{\theta}_1$ and $\boldsymbol{\theta}_2$ range over $[\frac{-2\pi}{3}, \frac{\pi}{6}]$, while for testing both inputs are sampled from $[-\pi, \pi]$.  Additionally, $\epsilon \sim \mathcal{N}(0, 0.1^2)$.

\subsubsection{Environmental Model 4D}
The Environmental Model function is a pollutant diffusion problem that models a pollutant spill at two locations caused by a chemical accident \cite{Bliznyuk2008BayesianCA}:

\begin{equation}
    \boldsymbol{y} = \sqrt{4\pi}\cdot C(X) + \epsilon,
\end{equation}
where $\epsilon \sim \mathcal{N}(0, 0.1^2)$, the response variable $\boldsymbol{y}$ is the scaled the concentration of the pollutant $C(X)$ at the space-time vector $\boldsymbol{(s, t)}$:
\begin{equation}
    C(X)  = \frac{\boldsymbol{M}}{\sqrt{4\pi \boldsymbol{D}t}} \exp \left( {\frac{-s^2}{4\boldsymbol{D}t}}\right) + C'(X)
\end{equation}
\begin{equation}
    C'(X) = \frac{\boldsymbol{M}}{\sqrt{4\pi \boldsymbol{D}(t-\boldsymbol{\tau})}} \exp \left( {\frac{-(s-\boldsymbol{L})^2}{4\boldsymbol{D}(t-\boldsymbol{\tau})}}\right)I(\boldsymbol{\tau}<t)
\end{equation}
where $\boldsymbol{M}$ is the mass of the pollutant spill, $\boldsymbol{D}$ is the diffusion rate in the channel, $\boldsymbol{L}$ is the location of the second spill, $\boldsymbol{\tau}$ is the time of the second spill and $I$ is the indicator function. For generating this dataset, $s$ and $t$ are fixed: $s = 1$ and $t = 40.1$. The ranges of the input values for training are as follows: $\boldsymbol{M} \in [7, 13], \boldsymbol{D} \in [0.02, 0.12], \boldsymbol{L} \in [0.01, 3], \boldsymbol{\tau} \in [30.01, 30.295]$. For testing, those ranges are wider: $\boldsymbol{M} \in [5, 15], \boldsymbol{D} \in [0, 0.15], \boldsymbol{L} \in [0.01, 3.2], \boldsymbol{\tau} \in [23.71, 31]$. 

\subsubsection{Planar arm torque 6D}
Planar arm torque dataset approximates the first motor’s torque in the inverse dynamics of a Planar 2D Arm \cite{cully2018limbo}:

\begin{equation}
    \boldsymbol{y}_{1} = (M(\boldsymbol{q})\ddot{\boldsymbol{q}} + C(\boldsymbol{q}, \dot{\boldsymbol{q}})\dot{\boldsymbol{q}})^T + \epsilon,
\end{equation}
where $\epsilon \sim \mathcal{N}(0, 0.1^2)$, $\boldsymbol{q}$ is a 2-dimensional vector denoting the articular position, $\dot{\boldsymbol{q}}$ is a 2-dimensional vector denoting the articular velocity, $\ddot{\boldsymbol{q}}$ is a 2-dimensional vector denoting the articular acceleration. $M(\boldsymbol{q})$ is the mass matrix and $C(\boldsymbol{q}, \dot{\boldsymbol{q}})$ is the matrix of Coriolis and centrifugal forces:

\begin{equation}
    M(\boldsymbol{q}) = \begin{bmatrix}
            0.2083 + 0.1250\cos(\boldsymbol{q}_{2}) & m(\boldsymbol{q}_{2}) \\
            m(\boldsymbol{q}_{2}) & 0.0417
            \end{bmatrix}
\end{equation}
\begin{equation}
     m(\boldsymbol{q}_{2}) = 0.0417 + 0.0625 \cos(\boldsymbol{q}_{2})
\end{equation}

\begin{equation}
\label{coriolis}
    C(\boldsymbol{q}, \dot{\boldsymbol{q}}) = \begin{bmatrix}
            - a\sin(\boldsymbol{q}_{2})\dot{\boldsymbol{q}}_{2} & a \sin(\boldsymbol{q}_{2})(\dot{\boldsymbol{q}}_{1}+\dot{\boldsymbol{q}}_{2})\\
            a \sin(\boldsymbol{q}_{2})\dot{\boldsymbol{q}}_{1} & 0
            \end{bmatrix},
\end{equation}
where $a = 0.0625$. The features span though the following intervals for training: $\boldsymbol{q}_1, \boldsymbol{q}_2 \in [-\frac{\pi}{2}, \frac{\pi}{2}]$, $\dot{\boldsymbol{q}}_1, \dot{\boldsymbol{q}}_2 \in [-\pi, \pi]$, $\ddot{\boldsymbol{q}}_1, \ddot{\boldsymbol{q}}_2 \in [-\pi, \pi]$. For generating the testing feature values $\boldsymbol{q}_1, \boldsymbol{q}_2 \in [-\pi, \pi]$, $\dot{\boldsymbol{q}}_1, \dot{\boldsymbol{q}}_2 \in [-2\pi, 2\pi]$, $\ddot{\boldsymbol{q}}_1, \ddot{\boldsymbol{q}}_2 \in [-2\pi, 2\pi]$ are used. 

\subsubsection{Piston simulation 7D}
The Piston Simulation function models the circular motion of a piston within a cylinder. The piston’s linear motion is transformed into circular motion by connecting a linear rod to a disk. Thus, the faster the piston moves inside the cylinder, the quicker the disk rotation and thus, the faster the engine runs. The response variable $\boldsymbol{y}$ is the cycle time in seconds \cite{BenAri2007ModelingDF, Freitas1999ModernIS}, which is affected by the features via a chain of nonlinear functions:

\begin{equation}
    \boldsymbol{y} = 2\pi\sqrt{\frac{\boldsymbol{M}}{\boldsymbol{k} + \boldsymbol{S}^2\frac{\boldsymbol{P}_{0} \boldsymbol{V}_{0}}{\boldsymbol{T}_{0}}\frac{\boldsymbol{T}_{a}}{\boldsymbol{V}^2}}} + \epsilon, \text{where}
\end{equation} 

\begin{equation}
    \boldsymbol{V} = \frac{\boldsymbol{S}}{2\boldsymbol{k}}\left(\sqrt{\boldsymbol{A}^2 + 4\boldsymbol{k}\frac{\boldsymbol{P}_{0}\boldsymbol{V}_{0}}{\boldsymbol{T}_{0}}\boldsymbol{T}_{a}} - \boldsymbol{A}   \right)
\end{equation}

\begin{equation}
    \boldsymbol{A} = \boldsymbol{P}_{0}\boldsymbol{qS} + 19.62\boldsymbol{M} - \frac{\boldsymbol{kV}_{0}}{\boldsymbol{S}}.
\end{equation}
In the above equations $\boldsymbol{M}$ is the piston weight (kg), $\boldsymbol{S}$ is the piston surface area ($\text{m}^2$), $\boldsymbol{V}_{0}$ is the initial gas volume ($\text{m}^3$), $\boldsymbol{k}$ is the spring coefficient (N/m), $\boldsymbol{P}_{0}$ is the atmospheric pressure (N/$\text{m}^2$), $\boldsymbol{T}_{a}$ is the ambient temperature (K), $\boldsymbol{T}_{0}$ is the filling gas temperature (K) and the error term $\epsilon \sim \mathcal{N}(0, 0.1^2)$. The training features are from the following intervals: $\boldsymbol{M} \in [30, 60]$, $\boldsymbol{S} \in [0.005, 0.020]$ , $\boldsymbol{V}_0 \in [0.002, 0.010]$, $\boldsymbol{k} \in [1000, 5000]$, $\boldsymbol{P}_0 \in [90000, 110000]$, $\boldsymbol{T}_a \in [290, 296]$ , $\boldsymbol{T}_0 \in [340, 360]$. Comparably, the testing input values are: $\boldsymbol{M} \in [0, 90]$, $\boldsymbol{S} \in [0.005, 0.03]$ , $\boldsymbol{V}_0 \in [0, 0.015]$, $\boldsymbol{k} \in [10, 6000]$, $\boldsymbol{P}_0 \in [80000, 120000]$, $\boldsymbol{T}_a \in [285, 300]$ , $\boldsymbol{T}_0 \in [300, 400]$.

\subsubsection{Robot arm 8D}
The Robot Arm function models the position of a four-segment robot arm and the response is the distance from the end of the robot arm to the origin \cite{An2001Quasiregression}:

\begin{equation}
    \boldsymbol{y} = \sqrt{\boldsymbol{u}^2 + \boldsymbol{v}^2} + \epsilon, \text{where}
\end{equation} 

\begin{equation}
    \boldsymbol{u} = \displaystyle\sum_{i=1}^{4}  \boldsymbol{L}_{i} \cos \left( \displaystyle\sum_{j=1}^{i} \boldsymbol{\theta}_{j} \right)
\end{equation}

\begin{equation}
    \boldsymbol{v} = \displaystyle\sum_{i=1}^{4}  \boldsymbol{L}_{i} \sin \left( \displaystyle\sum_{j=1}^{i} \boldsymbol{\theta}_{j} \right).
\end{equation}
The shoulder of the robot arm is fixed at the origin, however, each of the four segments is positioned at angle $\boldsymbol{\theta}_{j}$ and has length $\boldsymbol{L}_{i}$. Each input variable for the training set is generated from $\boldsymbol{\theta}_j \in [0, \pi]$ and $\boldsymbol{L}_i \in [0, 0.5]$, while the test input variables $\boldsymbol{\theta}_j$ and $\boldsymbol{L}_i$ range over $[0, 2\pi]$ and $[0, 1]$ respectively. Finally, $\epsilon \sim \mathcal{N}(0, 0.1^2)$.

\subsubsection{Borehole 8D}
The Borehole function models water flow through a borehole and thus, the response variable is the water flow rate ($\text{m}^3$/yr):

\begin{equation}
    \boldsymbol{y} = \frac{2\pi \boldsymbol{T}_{u}(\boldsymbol{H}_{u} - \boldsymbol{H}_{l})}{\text{ln}(\boldsymbol{r}/\boldsymbol{r}_{w})\left(1+\frac{2\boldsymbol{L}\boldsymbol{T}_{u}}{\text{ln}(\boldsymbol{r}/\boldsymbol{r}_{w})\boldsymbol{r}_{w}^2\boldsymbol{K}_{w}} + \frac{\boldsymbol{T}_{u}}{\boldsymbol{T}_{l}} \right)} + \epsilon,
\end{equation}
where $\epsilon \sim \mathcal{N}(0, 0.1^2)$, $\boldsymbol{r}_w$ is the radius of a borehole (m), $\boldsymbol{r}$ is the radius of influence (m), $\boldsymbol{T}_u$ and $\boldsymbol{T}_l$ are the transmissivities of respectively upper and lower aquifers ($\text{m}^2$/yr), $\boldsymbol{H}_u$ and $\boldsymbol{H}_l$ are the potentiometric heads of respectively upper and lower aquifers (m), $\boldsymbol{L}$ is the length of a borehole (m) and $\boldsymbol{K}_w$ is the hydraulic conductivity of borehole (m/yr) \cite{An2001Quasiregression}. The features for training are sampled from $\boldsymbol{r}_w \in [0.05, 0.15]$, $\boldsymbol{r} \in [100, 50000]$, $\boldsymbol{T}_u \in [63070, 115600]$, $\boldsymbol{T}_l \in [63.1, 116]$, $\boldsymbol{H}_u \in [990, 1110]$, $\boldsymbol{H}_l \in [700, 820]$, $\boldsymbol{L} \in [1120, 1680]$, $\boldsymbol{K}_w \in [9855, 12045]$, while the testing input - $\boldsymbol{r}_w \in [0.01, 0.2]$, $\boldsymbol{r} \in [90, 50010]$, $\boldsymbol{T}_u \in [63020, 115650]$, $\boldsymbol{T}_l \in [60, 120]$, $\boldsymbol{H}_u \in [950, 1150]$, $\boldsymbol{H}_l \in [650, 900]$, $\boldsymbol{L} \in [1100, 1700]$, $\boldsymbol{K}_w \in [9800, 12100]$. 

\subsubsection{PUMA560 9D}
PUMA560 dataset is generated from a realistic simulation of the dynamics of a Puma 560 robot arm \cite{Ghahramani1996Pumadyn}. The task is to predict the articular acceleration of one of the links of the robot arm:

\begin{equation}
    \boldsymbol{y}_{1} = A(\boldsymbol{q})^{-1}(\boldsymbol{\tau} - n(\boldsymbol{q}, \dot{\boldsymbol{q}}) - g(\boldsymbol{q})) + \epsilon,
\end{equation}
where $\boldsymbol{q}$ and $\dot{\boldsymbol{q}}$ are 3-dimensional vectors denoting respectively the angular positions and angular velocities of each of the three links, $\boldsymbol{\tau}$ is a 3-dimensional vector representing the torques at the three joints, $n(\boldsymbol{q}, \dot{\boldsymbol{q}})$ is the Coriolis and centrifugal effects, $A(\boldsymbol{q})$ is the inertia matrix, $g$ is the gravity and $\epsilon \sim \mathcal{N}(0, 0.4^2)$ is the Gaussian noise. The test and train features are sampled from $\boldsymbol{q}_1, \boldsymbol{q}_2, \boldsymbol{q}_3 \in \beta[\frac{-\pi}{2}, \frac{\pi}{2}]$, $\dot{\boldsymbol{q}}_1, \dot{\boldsymbol{q}}_2, \dot{\boldsymbol{q}}_3 \in \beta[\frac{-\pi}{2}, \frac{\pi}{2}]$ and $\boldsymbol{\tau}_1, \boldsymbol{\tau}_2, \boldsymbol{\tau}_3 \in \beta[\frac{-1}{2}, \frac{1}{2}]$ with fixed $\beta =1.2$ to control the nonlinearity of the dataset. Therefore, this dataset can be considered as highly nonlinear and noisy.

\subsubsection{Wing weight 10D}
The Wing Weight function models a light aircraft wing, where the response is the wing's weight  \cite{Forrester2008EngineeringDV}:

\begin{equation}
    \boldsymbol{y} = 0.036\boldsymbol{S}_{w}^{0.758}\boldsymbol{W}_{fw}^{0.0035}\left( \frac{\boldsymbol{A}}{\cos^2(\boldsymbol{\Lambda})} \right)^{0.6}\boldsymbol{q}^{0.006}\boldsymbol{y}'
\end{equation}
\begin{equation}
    \boldsymbol{y}' = \boldsymbol{\lambda}^{0.04}\left( \frac{100\boldsymbol{t}_{c}}{\cos(\boldsymbol{\Lambda})} \right)^{-0.3}(\boldsymbol{N}_{z}\boldsymbol{W}_{dg})^{0.49} + \boldsymbol{y}''
\end{equation}
\begin{equation}
    \boldsymbol{y}'' = \boldsymbol{S}_{w}\boldsymbol{W}_{p} + \epsilon,
\end{equation}
where $\boldsymbol{S}_{w}$ is the wing area ($\text{ft}^2$), $\boldsymbol{W}_{fw}$ is the weight of fuel in the wing (lb), $\boldsymbol{A}$ is the aspect ratio, $\boldsymbol{\Lambda}$ is the quarter-chord sweep (degrees), $\boldsymbol{q}$ is the dynamic pressure at cruise (lb/$\text{ft}^2$), $\boldsymbol{\lambda}$ is the taper ratio, $\boldsymbol{t}_{c}$ is the aerofoil thickness to chord ratio, $\boldsymbol{N}_{z}$ is the ultimate load factor, $\boldsymbol{W}_{dg}$ is the flight design gross weight (lb), $\boldsymbol{W}_{p}$ is the paint weight (lb/$\text{ft}^2$) and $\epsilon \sim \mathcal{N}(0, 0.1^2)$. The ranges of the features for training for each value are $\boldsymbol{S}_w \in [150, 200]$, $\boldsymbol{W}_{fw}  \in [220, 300]$, $\boldsymbol{A}  \in [6, 10]$, $\boldsymbol{\Lambda}  \in [-10, 10]$, $\boldsymbol{q} \in [16, 45]$, $\boldsymbol{\lambda} \in [0.5, 1]$, $\boldsymbol{t}_c  \in [0.08, 0.18]$, $\boldsymbol{N}_z  \in [2.5, 6]$, $\boldsymbol{W}_{dg} \in [1700, 2500]$ and $\boldsymbol{W}_p \in [0.025, 0.08]$, whereas for testing the intervals contain values outside the usual ranges - $\boldsymbol{S}_w \in [100, 250]$, $\boldsymbol{W}_{fw}  \in [200, 320]$, $\boldsymbol{A}  \in [0, 15]$, $\boldsymbol{\Lambda}  \in [-20, 20]$, $\boldsymbol{q} \in [0, 60]$, $\boldsymbol{\lambda} \in [0, 1.5]$, $\boldsymbol{t}_c  \in [0.05, 0.25]$, $\boldsymbol{N}_z  \in [0.5, 8]$, $\boldsymbol{W}_{dg} \in [1000, 3000]$ and $\boldsymbol{W}_p \in [0, 0.1]$. 

\subsection{Many local minima functions}
The many local minima functions are extremely hard to be approximated due to their high nonlinearity and complexity. 

\subsubsection{Schaffer N.4 2D}
Schaffer N.4 function, proposed in \cite{Mishra2006SomeNT}:

\begin{equation}
    \boldsymbol{y} = 0.5 + \frac{\cos^2(\sin(|\boldsymbol{x}_{1}^2 - \boldsymbol{x}_{2}^2)) - 0.5}{(1+0.001(\boldsymbol{x}_{1}^2 + \boldsymbol{x}_{2}^2))^2} + \epsilon,
\end{equation}
where $\epsilon$ is the added noise and $\epsilon \sim \mathcal{N}(0, 0.1^2)$. The training inputs $\boldsymbol{x}_1$ and $\boldsymbol{x}_2$ are generated from $[-2,2]$, while for testing $\boldsymbol{x}_1$ and $\boldsymbol{x}_2$ range over $[-2.5,2.5]$.

\subsubsection{Rastrigin 3D}
The Rastrigin function is highly multimodal, but locations of the minima are regularly distributed. Rastrigin proposed the function in two dimensions \cite{rastrigin1974systems}, which was later generalized to $d$ dimensions by Rudolph \cite{rudolph1990globale}. However, for the purpose of this work, the function is used in 3 dimensions ($d=3$):

\begin{equation}
    \boldsymbol{y} = 10d + \displaystyle\sum_{i=1}^{d}(\boldsymbol{x}_{i}^2 - 10\cos(2\pi \boldsymbol{x}_{i})) + \epsilon,
\end{equation}
where $\epsilon \sim \mathcal{N}(0, 0.1^2)$. Each training features $\boldsymbol{x}$ is sampled from $[-5.12, 5.12]$, whereas the testing features are generated from $[-5.5, 5.5]$.

\subsubsection{Griewank 4D}
The local minima of the Griewank function are widespread and regularly distributed. It is presented in $d$ dimensions   \cite{Griewank1981GeneralizedDF}:

\begin{equation}
    \boldsymbol{y} =  \displaystyle\sum_{i=1}^{d} \frac{\boldsymbol{x}_{i}^2}{4000} - \displaystyle\prod_{i=1}^d\cos\left(\frac{\boldsymbol{x}_{i}}{\sqrt{i}}\right) + 1 + \epsilon,
\end{equation}
where $\epsilon \sim \mathcal{N}(0, 0.1^2)$ is homoscedastic noise. The dimenionality of this function chosen for the aims of this work is $d=4$. The training features $\boldsymbol{x}$ are generated from $[-500, 500]$, while the testing feature vectors from $[-600, 600]$.

\subsubsection{Ackley 7D} 
Ackley function is introduced as a $d$-dimensional function \cite{Ackley1987ACM}:

\begin{equation}
    \boldsymbol{y} = -a\exp\left( -b \sqrt{\frac{1}{d}\displaystyle\sum_{i=1}^{d}\boldsymbol{x}_{i}^2} \right) - \boldsymbol{y}'
\end{equation}
\begin{equation}
    \boldsymbol{y}' = \exp\left( \frac{1}{d}\displaystyle\sum_{i=1}^{d}\cos(c\boldsymbol{x}_{i}) \right) + a + \exp(1) + \epsilon,
\end{equation}
where $a = 20, b = 0.2, c = 2\pi$ and $\epsilon \sim \mathcal{N}(0, 0.1^2)$. For the purposes of this work, $d = 7$, $\boldsymbol{x}$ for training are sampled from $[-30, 30]$ and $\boldsymbol{x}$ for testing are generated from $[-32.768, 32.768]$, as the latter is the usual test area \cite{Molga2005TestftionsforOptim}.

\subsection{Trigonometric}
\label{trigapp}
\subsubsection{He et al. 1D} 
\label{expHe}
\begin{figure}[t]
\begin{subfigure}{.23\textwidth}
  \centering
  \includegraphics[width=\linewidth]{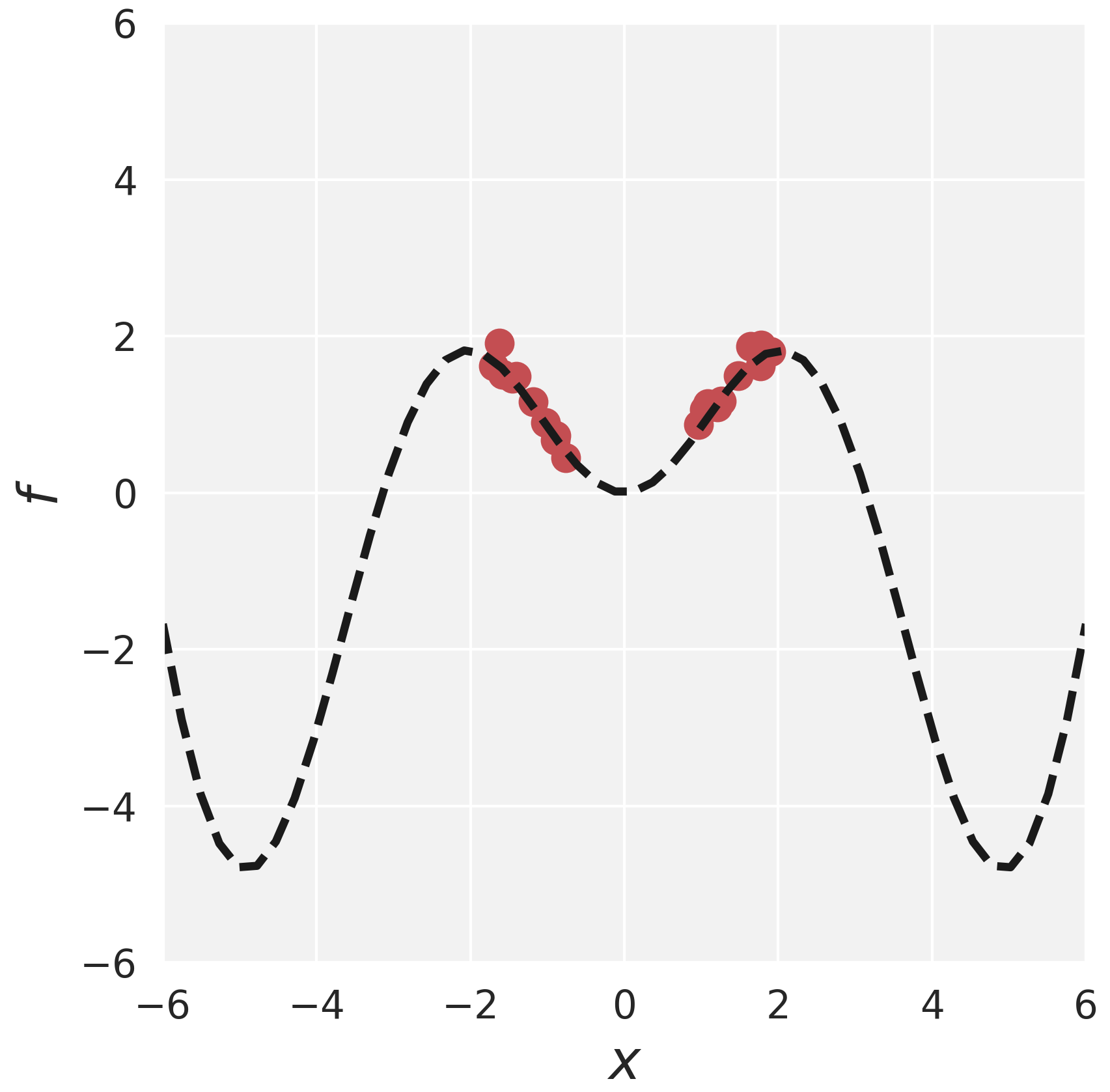}
  \caption{He et al.}
  \label{fig:xsinxTA}
\end{subfigure}%
\vspace{1mm}
\begin{subfigure}{.23\textwidth}
  \centering
  \includegraphics[width=\linewidth]{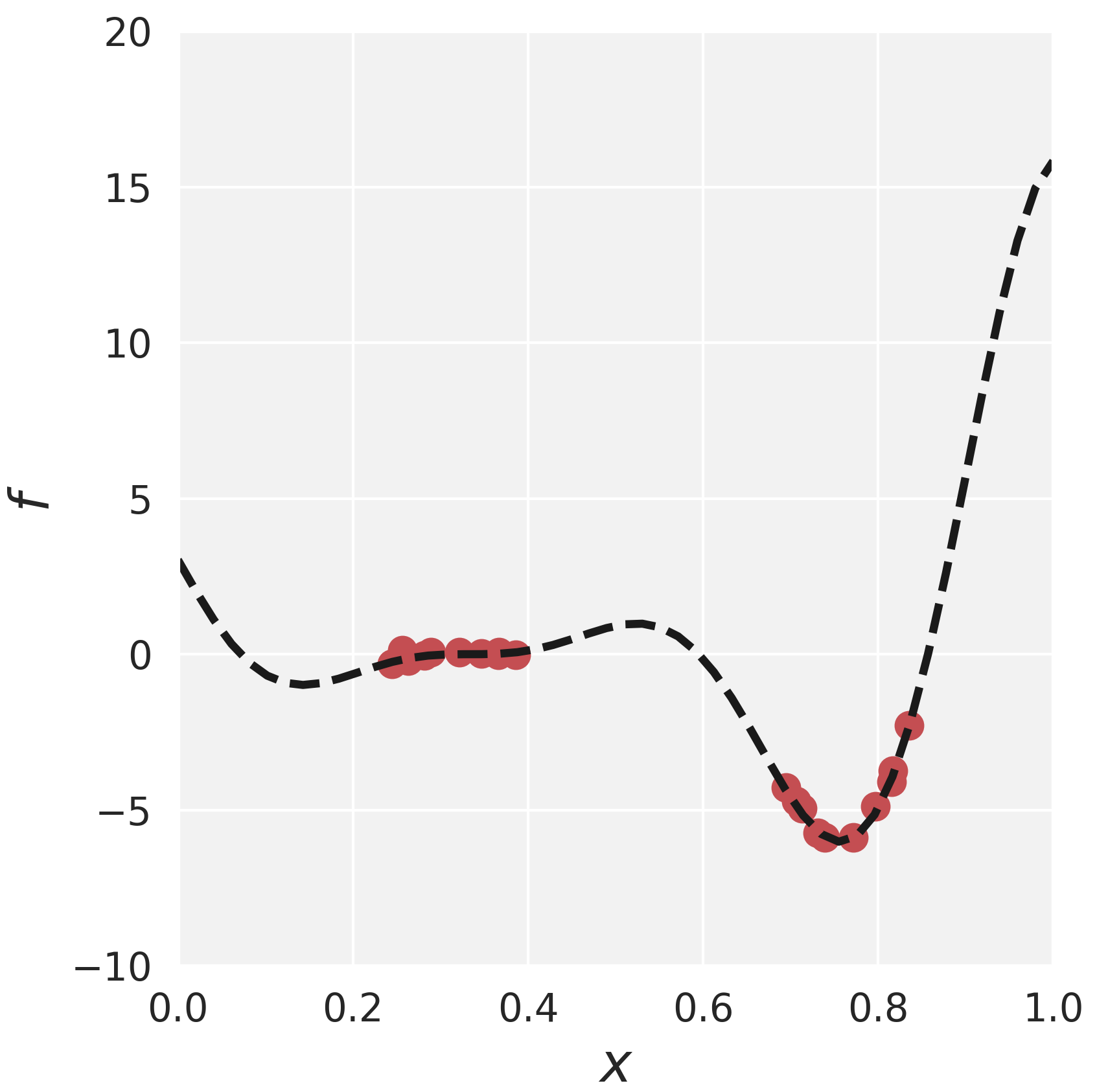}
  \caption{Forrester et al.}%
  \label{fig:forresterTA}
\end{subfigure}
\caption{Training data points for the two 1D generating functions.}
\label{fig:datasets1D}
\end{figure}

The generating function:
\begin{equation}
    \boldsymbol{y} = \boldsymbol{x}\sin(\boldsymbol{x}) + \epsilon,
\end{equation}
where $\epsilon \sim \mathcal{N}(0, 0.1^2)$, is proposed by He et al. \cite{He2020BayesianDE}. In order to detail uncertainty on out-of-distribution test data, the training points are partitioned into two equal-sized clusters (Figure \ref{fig:xsinxTA}). The clusters, that configure the training data $\boldsymbol{x}$, are generated from $[-2, -0.67]$ and $[0.67, 2]$, while the testing points $\boldsymbol{x}$ range in $[-6,6]$.


\subsubsection{Forrester et al. 1D}
\label{expFor}
Forrester et al. function is a simple one-dimensional multimodal function    \cite{Forrester2008EngineeringDV}:

\begin{equation}
    \boldsymbol{y} = (6\boldsymbol{x}-2)^2\sin(12\boldsymbol{x}-4) + \epsilon,
\end{equation}
where $\epsilon \sim \mathcal{N}(0, 0.1^2)$. Similarly to He et al. dataset, training vector $\boldsymbol{x}$ is split into two clusters to detail uncertainty on OoD test data (Figure  \ref{fig:forresterTA}). Thus, $\boldsymbol{x}$ ranges over $[0.2, 0.4]$ and $[0.65, 0.85]$, while testing $\boldsymbol{x}$ is sampled from $[0, 1]$.

\subsubsection{Ishigami 3D}
Ishigami function, introduced in \cite{Ishigami1990AnIQ}, shows strong non-linearity and non-monotonicity as well as characteristic dependence on $x_3$ \cite{Sobol1999OnTU}:

\begin{equation}
    \boldsymbol{y} = \sin(\boldsymbol{x}_{1}) + a\sin^2(\boldsymbol{x}_{2}) + b \boldsymbol{x}_{3}^4\sin(\boldsymbol{x}_{1}) + \epsilon,
\end{equation}
where $a = 7$ and $b = 0.1$, following \cite{crestaux2009}. Also, $\epsilon \sim \mathcal{N}(0, 0.1^2)$. The training values of $\boldsymbol{x}_1, \boldsymbol{x}_2, \boldsymbol{x}_3$ are sampled from $[-\frac{\pi}{2}, \frac{\pi}{2}]$. Similarly,  $\boldsymbol{x}_1, \boldsymbol{x}_2, \boldsymbol{x}_3$ for testing are sampled $[-\frac{2\pi}{3}, \frac{2\pi}{3}]$ respectively.

\subsubsection{Friedman 5D}
Friedman et al. have proposed the following five-dimensional function   \cite{Freidman1991MultivariateAR, Friedman1983MultidimensionalAS}:

\begin{equation}
    \boldsymbol{y} = 10\sin(\pi \boldsymbol{x}_{1}\boldsymbol{x}_{2}) + 20(\boldsymbol{x}_{3}-0.5)^2 + 10\boldsymbol{x}_{4} +5\boldsymbol{x}_{5} + \epsilon,
\end{equation}
where $\epsilon \sim \mathcal{N}(0, 0.1^2)$. The training data $\boldsymbol{x}_1, \boldsymbol{x}_2, \boldsymbol{x}_3, \boldsymbol{x}_4$ and $\boldsymbol{x}_5$ are sampled from $[0, 0.5]$, while the testing data $\boldsymbol{x}_1, \boldsymbol{x}_2, \boldsymbol{x}_3, \boldsymbol{x}_4$ and $\boldsymbol{x}_5$ is generated from $[0, 1]$.\\

\subsection{Others}

\subsubsection{Roos \& Arnold 5D}
The Roos \& Arnold function, proposed in \cite{Roos1963NumerischeEZ}, is formed from the products of one-dimensional functions:

\begin{equation}
    \boldsymbol{y} = \displaystyle\prod_{i=1}^{d}|4\boldsymbol{x}_{i} - 2| + \epsilon,
\end{equation}
where $\epsilon \sim \mathcal{N}(0, 0.1^2)$ and $d=5$. It is described by Kucherenko et al. as a function with dominant high-order interaction terms and a high effective dimension \cite{Kucherenko2011TheIO}. $\boldsymbol{x}$ for training  and $\boldsymbol{x}$ for testing are sampled respectively from $[0, 0.8]$ and $[0, 1]$.\\

\subsubsection{Sum of powers 6D}
This bowl-shaped $D$-dimensional function, introduced in \cite{Molga2005TestftionsforOptim}, represents a sum of different powers:

\begin{equation}
    \boldsymbol{y} = \displaystyle\prod_{i=1}^{d}|\boldsymbol{x}_{i}|^{i+1} + \epsilon,
\end{equation}
where $\epsilon \sim \mathcal{N}(0, 0.1^2)$ and $d=6$. The training data $\boldsymbol{x}_i$ ranges over $[-0.75, 0.75]$, while the testing data $\boldsymbol{x}_i$ from $[-1, 1]$.\\

\subsubsection{Styblinski-Tang 9D} 
Styblinski-Tang function is proposed as a function in $d$ dimensions \cite{Yi2020AnEO}:

\begin{equation}
    \boldsymbol{y} = \frac{1}{2}\displaystyle\sum_{i=1}^{d}(\boldsymbol{x}_{i}^4 - 16\boldsymbol{x}_{i}^2 + 5\boldsymbol{x}_{i}) + \epsilon,
\end{equation}
where $\epsilon \sim \mathcal{N}(0, 0.1^2)$ and $d=9$. The testing  $\boldsymbol{x}_i$ and training $\boldsymbol{x}_i$ feature vectors are sampled from the intervals $[-5, 5]$ and $[-6, 6]$ respectively. \\

\subsubsection{Adapted Welch et al. 10D}
The original function, proposed by Welch et al. \cite{Welch1992ScreeningPA}, contains 20 dimensions such that some input variables have a very high effect on the output, compared to others. This function is considered challenging, because of its interactions and nonlinear effects. To fit the needs of this work, the Welch et al. function is adapted and its new version has 10 dimension, while still preserving its characteristics:

\begin{equation}
    \boldsymbol{y} =  \frac{5\boldsymbol{x}_{10}}{1.001+\boldsymbol{x}_{1}} + 5(\boldsymbol{x}_{4}-\boldsymbol{x}_{2})^2 + \boldsymbol{x}_{5} + 40\boldsymbol{x}_{9}^3 + \boldsymbol{y}'
\end{equation}
\begin{equation}
    \boldsymbol{y}' =  - 5\boldsymbol{x}_{1} + 0.08\boldsymbol{x}_{3} + 0.25\boldsymbol{x}_{6}^2 + \boldsymbol{y}''
\end{equation}
\begin{equation}
    \boldsymbol{y}'' = 0.03\boldsymbol{x}_{7} - 0.09\boldsymbol{x}_{8} + \epsilon,
\end{equation}
where $\epsilon \sim \mathcal{N}(0, 0.1^2)$. The ranges of training features $\boldsymbol{x}_1, \boldsymbol{x}_2, \boldsymbol{x}_3, \boldsymbol{x}_4, \boldsymbol{x}_5, \boldsymbol{x}_6, \boldsymbol{x}_7, \boldsymbol{x}_8, \boldsymbol{x}_9$ and $\boldsymbol{x}_{10}$ are $[-0.5, 0.5]$. The testing features $\boldsymbol{x}_1, \boldsymbol{x}_2, \boldsymbol{x}_3, \boldsymbol{x}_4, \boldsymbol{x}_5, \boldsymbol{x}_6, \boldsymbol{x}_7, \boldsymbol{x}_8, \boldsymbol{x}_9$ and $\boldsymbol{x}_{10}$ and $[-1, 1]$. 

\section{Real-world datasets}
\label{real-worlddata}
The number of training data points and testing data points for each dataset is shown in Table \ref{numberpoints}.
\subsection{Boston housing}
The goal of the Boston housing dataset is to predict the price of a house given its number of rooms and other context factors. However, in this paper, the number of rooms is the only considered independent variable and context factors, such as house location and crime rate by town, are disregarded. \\

\subsection{Abalone shells}
The Abalone dataset contains data regarding abalone shells and the regression task is to predict the age of a shell, determined by the number of rings, given several physical measurements  \cite{Abalone1994}. The features used in this study are: length (denoting the longest shell measurement in mm), diameter in mm, height (with meat in shell  in mm), whole weight (whole abalone in grams) and sucked weight (meat weight in grams). \\

\subsection{Naval propulsion plant}
The naval propulsion plant dataset has been generated from a sophisticated simulator of a Gas Turbine (GT) \cite{Coraddu2013Machine}. The task is to predict the GT propulsion plant’s decay state coefficient. Originally, the features are given in a 16-dimensional feature vector containing the GT measures at steady state of the physical asset, but in this work only GT shaft torque (kN/m), GT rate of revolutions (rpm), high pressure turbine exit temperature (C) and GT exhaust gas pressure (bar),  are being used as features. \\

\subsection{Forest fire}
This dataset concerns forest fire data from the Montesinho natural park of Portugal \cite{Cortez2007ADM}. The aim of this dataset is to predict the burnt area given Fine Fuel Moisture Cod (FFMC) index, Duff Moisture Code (DMC) index, Drought Code (DC) index, Initial Spread (ISI) index, temperature in Celsius degrees and relative humidity (in percentage). The full set of attributes of this data set includes also spatial coordinates within the park, day and month, wind and speed, but those are discarded as their addition provides too detailed context information contradicting this study’s goals. Additionally, the dependent variable was first transformed with a $\ln(x+1)$ function, just like in \cite{Cortez2007ADM}. \\

\subsection{Parkinson's disease} 
Parkinson’s disease dataset is composed of a range of biomedical voice measurements from people with Parkinson's disease (PD) \cite{Little2007ExploitingNR}. In total there are 23 features, but only five are being used: NHR and HNR, which are both measures of ratio of noise to tonal components in the voice status, DFA, which is a signal fractal scaling exponent, PPE, denoting three nonlinear measures of fundamental frequency variation and RPDE, which is a nonlinear dynamical complexity measure. Therefore, the goal of this regression task it to predict the total Unified Parkinson’s Disease Rating Scale (UPDRS) given the aforementioned five features. \\\\

\begin{table}[t!]
\caption{Number of training data points and testing data points for each dataset. \label{numberpoints}}
 \centering
{\small
\begin{tabular}{||l c c||}
\hline
 & Training & Testing \\[0.2ex] 
 \hline\hline
 
 He et al. 1D  & 20  & 50\\

 Forrester et al. 1D   &  20 &  50\\

 Schaffer N.4 2D   & 1000 & 2500\\
 
 Double pendulum 2D   & 1000 & 2500\\
 
 Rastrigin 3D & 200 &  500\\
 
 Ishigami 3D   &  2000 & 5000 \\
 
 Environmental 4D & 200 & 500\\
 
 Griewank 4D   & 200 & 500\\
 
 Roos \& Arnold 5D   & 200 & 500 \\
 
 Friedman 5D  & 200 & 500\\
 
 Planar arm torque 6D   & 200 & 500\\
 
 Sum of powers 6D  & 200 & 500\\
 
 Ackley 7D  & 400 & 1000\\
 
 Piston simulation 7D  & 200 & 500\\
 
 Robot arm 8D   & 200 & 500 \\
 
 Borehole 8D    &  2000 & 5000\\
 
 Styblinski-Tang 9D  & 400  & 1000\\
 
 PUMA560 9D    &  3693  & 4499 \\
 
 Adapted Welch 10D  & 200  & 500 \\
 
 Wing weight 10D  & 2000  &  5000 \\
 
 Boston housing & 354  & 152\\
 
 Abalone &  1880 & 2297 \\
 
 Naval propulsion plant &  5370 & 6564\\
 
 Forest fire & 200 & 317\\
 
 Parkinson's  & 2643 & 3232\\
 
\hline
\end{tabular}}
\end{table}

\end{document}